\documentclass[10pt,journal,compsoc]{IEEEtran}
\usepackage{amsmath}
\usepackage{amssymb}
\usepackage[colorlinks,linkcolor=red,anchorcolor=blue,citecolor=green]{hyperref}
\usepackage[numbers]{natbib}
\usepackage{multirow}
\usepackage{algorithm}
\usepackage{algorithmic}

\usepackage{booktabs} 
\usepackage[pdftex]{graphicx}
\usepackage[table]{xcolor}
\usepackage{tabularx}
\ifCLASSINFOpdf
\usepackage[pdftex]{graphicx}
\DeclareGraphicsExtensions{.pdf,.jpeg,.png}
\else
\fi
\usepackage{placeins}

\usepackage{amsthm}
\newtheorem{Property}{Property}

\begin{document}

\title{Graph Generation with Spectral Geodesic Flow Matching}

\author{Xikun Huang,
        Tianyu Ruan,
        Chihao Zhang,
        Shihua Zhang
\IEEEcompsocitemizethanks{\IEEEcompsocthanksitem Xikun Huang, Tianyu Ruan, Chihao Zhang and Shihua Zhang* are with the State Key Laboratory of Mathematical Sciences, Academy of Mathematics and Systems Science, Chinese Academy of Sciences, Beijing 100190, China, and School of Mathematics Sciences, University of Chinese Academy of Sciences, Beijing 100049, China. Shihua Zhang is also with the Key Laboratory of Systems Health Science of Zhejiang Province, School of Life Science, Hangzhou Institute for Advanced Study, University of Chinese Academy of Sciences, Chinese Academy of Sciences, Hangzhou 310024, China\\
Xikun Huang and Tianyu Ruan contributed equally to this work.\\
*To whom correspondence should be addressed. Email: zsh@amss.ac.cn.}% <-this % stops an unwanted space
}        

%\IEEEcompsocitemizethanks{\IEEEcompsocthanksitem M. Shell was with the Department of Electrical and Computer Engineering, Georgia Institute of Technology, Atlanta, GA, 30332.\protect\\

%E-mail: see http://www.michaelshell.org/contact.html \IEEEcompsocthanksitem J. Doe and J. Doe are with Anonymous University.}% <-this % stops an unwanted space


%\markboth{IEEE TRANSACTIONS ON NEURAL NETWORKS AND LEARNING SYSTEMS, VOL. x, NO. x, JANUARY 2025}%
%{Shell \MakeLowercase{\textit{et al.}}: Bare Demo of IEEEtran.cls for Computer Society Journals}

\IEEEtitleabstractindextext{%
\begin{abstract}
Graph generation is a fundamental task with wide applications in modeling complex systems. Although existing methods align the spectrum or degree profile of the target graph, they often ignore the geometry induced by eigenvectors and the global structure of the graph. In this work, we propose Spectral Geodesic Flow Matching (SFMG), a novel framework that uses spectral eigenmaps to embed both input and target graphs into continuous Riemannian manifolds. We then define geodesic flows between embeddings and match distributions along these flows to generate output graphs. Our method yields several advantages: (i) captures geometric structure beyond eigenvalues, (ii) supports flexible generation of diverse graphs, and (iii) scales efficiently. Empirically, SFMG matches the performance of state-of-the-art approaches on graphlet, degree, and spectral metrics across diverse benchmarks. In particular, it achieves up to 30× speedup over diffusion-based models, offering a substantial advantage in scalability and training efficiency. We also demonstrate its ability to generalize to unseen graph scales. Overall, SFMG provides a new approach to graph synthesis by integrating spectral geometry with flow matching.
\end{abstract}

}

\maketitle

\IEEEdisplaynontitleabstractindextext

\IEEEpeerreviewmaketitle

\IEEEraisesectionheading{\section{Introduction}
\label{introduction}}
%%%%%%%%% BODY TEXT
\IEEEPARstart{G}{}
  raphs are powerful tools for representing and analyzing relationships between entities in various domains. Graph generation is a critical yet challenging problem with broad applications such as combinatorial optimization \cite{sun2023difusco}, drug discovery \cite{li2018multi}, and inverse protein folding \cite{yi2023graph}. This problem has garnered significant attention, leading to the development of various methods.

  Deep learning-based graph generation methods are often inspired by well-known generative models designed for regular data structures, such as images or sequences, to the irregular structure of graphs. For example, GraphVAE \cite{simonovsky2018graphvae} extends variational autoencoders (VAEs) \cite{kingma2013auto} by jointly encoding nodes and adjacency matrices into a continuous latent space, allowing the model to generate new graphs in one shot. However, the method incurs high computational costs when applied to larger graphs and struggles to guarantee the validity of generated graphs. In contrast, GraphRNN \cite{you2018graphrnn} generates graphs autoregressively in an efficient manner. However, this approach is sensitive to permutations and prone to overfitting. SPECTRE \cite{martinkus2022}, on the other hand, built on generative adversarial networks (GANs) \cite{goodfellow2020generative}, faces the same training difficulties commonly associated with GANs. 
  
  Recently, diffusion models \cite{ho2020denoising,song2020denoising} have achieved significant success. These models utilize neural networks to learn and simulate reverse-time stochastic differential equations (SDEs) \cite{song2020score} and transform white noise into the data distribution. Diffusion models demonstrate the advanced capability to model complex dependencies and learn distributions. Several methods have been developed based on this. 
  For example, GDSS \cite{jo2022score} learns graph structures by modeling the distribution of the adjacency matrices. 
  However, directly adding isotropic Gaussian noise to the adjacency matrix is suboptimal, as it violates the discrete nature of adjacency matrices and may lead to invalid graph structures. To address this issue, GSDM \citep{luo2023fast} performs an eigendecomposition of the adjacency matrix and employs a diffusion model to learn the distribution of the eigenvalues while ignoring the eigenvectors. This approach achieves high generation quality with low computational cost. However, the omission of eigenvectors limits its ability to capture the graph structure fully. GGSD \citep{minellogenerating} extends this framework by applying a diffusion model to learn the distribution of eigenvectors in Euclidean space, but it neglects the intrinsic geometric constraints. Modeling the distribution of eigenvectors remains a fundamental challenge due to the orthogonal constraints, which render standard diffusion models inapplicable. 
  
  As an alternative to diffusion models, flow matching models \cite{lipman2023flow,albergo2023building, liu2023flow} transform white noise into data distributions via ordinary differential equations (ODEs). These models offer comparable performance to diffusion models while being more efficient and easy to implement, especially in geometric applications. Given these advantages, flow matching emerges as a promising framework for graph generation. However, implementing these advanced methods for graph generation is nontrivial due to several inherent challenges:
  \begin{enumerate}
  \item[$\bullet$] \textbf{Sparse topological structure}: Graph topological structures are represented by adjacency matrices, which are often sparse. However, generative models, mainly designed for Euclidean space and continuous modeling, frequently struggle to capture this sparse structure, potentially leading to invalid graphs.
  
  \item[$\bullet$] \textbf{Non-uniqueness of graph representations}: Graphs admit multiple representations, such as adjacency matrices, Laplacian matrix decompositions, or hyperbolic space embeddings. Different permutations in the adjacency matrix yield diverse representations, complicating the learning process.
  
  \item[$\bullet$]  \textbf{Learning distribution in representation space}: Graph learning faces the inherent challenge of limited data, especially compared to domains like images and sequences. 
  This drives the need for models that efficiently model distributions in high-dimensional spaces. 
  \end{enumerate}
  
\begin{figure*}[htb]
  \vskip 0.0in
  \begin{center}
  \centerline{\includegraphics[width=1.04\textwidth]{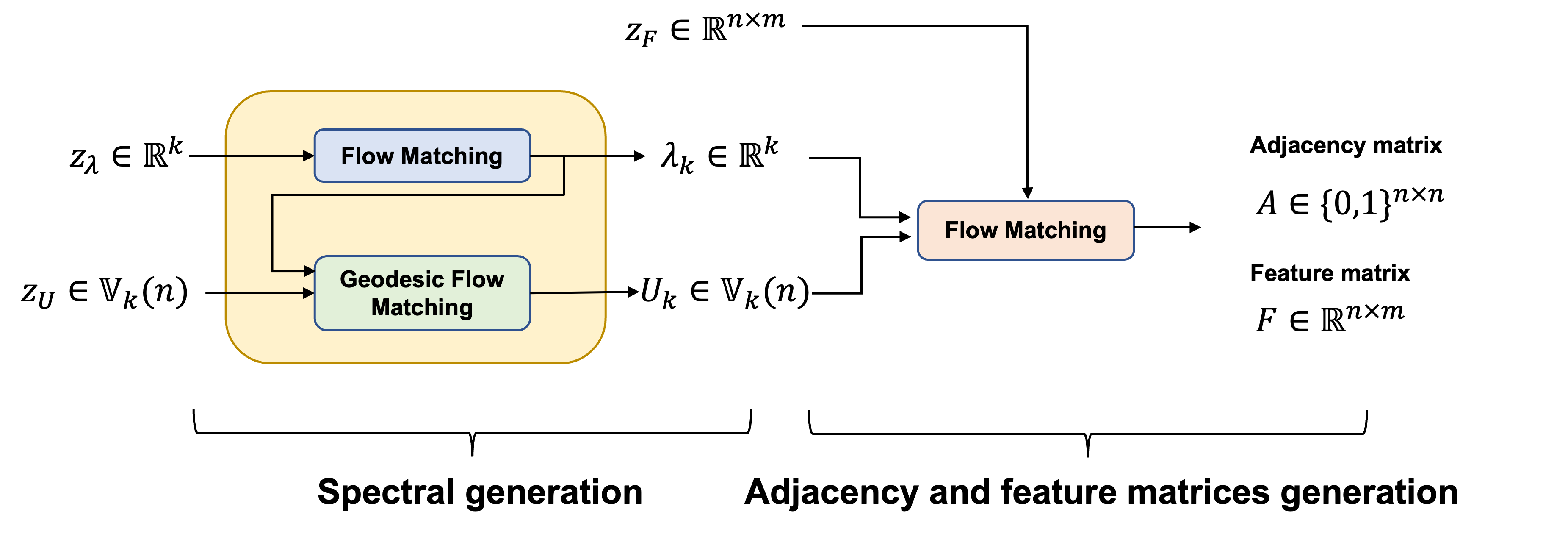}}
  \caption{Illustration of SFMG. SFMG adopts flow matching in $R^{k}$ and geodesic flow matching in $\mathbb{V}_k(n)$ to the first $k$ eigenvalues and eigenvectors. The last flow matching is used to recover the adjacency matrix from the generated Laplacian matrix and generate the feature matrix.}
  \label{fig:pipeline}
  \end{center}
  \vskip -0.3in
  \end{figure*}

  Addressing the first two challenges requires developing natural and effective representations for discrete graph structures. The third challenge demands efficient algorithms to model distributions in the chosen representation space. Here, we propose a spectral flow matching model for graph generation (SFMG). SFMG leverages spectral decomposition to derive compact graph representations, accommodating sparsity and resolving the non-uniqueness of graph representations by focusing on eigenvalues and eigenvectors. Furthermore, SFMG employs flow matching to enhance the effectiveness of distribution learning. It fully utilizes the geometric structure of the spectral space, thereby overcoming the challenges posed by limited graph data. This also offers superior generation efficiency compared to diffusion models. The key contribution of SFMG is its ability to model both eigenvalue and eigenvector distributions with eigenvectors modeled directly on the Stiefel manifold. To our knowledge, SFMG is the first method to achieve this. SFMG provides a more complete and expressive representation of graph structures by learning eigenvector distributions on manifolds. SFMG consists of three flow-matching models designed to generate (1) the first $k$ eigenvalues, (2) their corresponding eigenvectors, and (3) reconstruct the adjacency matrix from spectral information while generating node features (Figure \ref{fig:pipeline}).
  
  We evaluate SFMG on seven diverse graph datasets and compare it against state-of-the-art graph generative models. Extensive experimental results demonstrate that SFMG achieves high-quality learning of Laplacian eigenvalues and eigenvectors. In terms of overall graph generation, our method shows superior performance to existing approaches across different graph scales and applications as measured by diverse metrics. Notably, SFMG achieves a 30× speedup over diffusion-based methods—avoiding their scalability issues on large graphs—while matching state-of-the-art autoregressive models in structural fidelity.
  
  \section{Related Work}
  \textbf{Graph Generative Models}.
  Graph generative models are typically categorized as autoregressive or one-shot. Autoregressive models \cite{you2018graphrnn,liao2019} construct graphs sequentially by adding nodes and edges, allowing for constraint integration but suffering from limitations related to node ordering. In contrast, one-shot models generate the entire graph in a single step, naturally supporting permutation invariance or equivariance. These include variational autoencoders \cite{simonovsky2018graphvae}, GANs \cite{martinkus2022}, normalizing flows \cite{luo2021graphdf}, and diffusion models \cite{niu2020, jo2022score, luo2023fast, vignac2023}, with diffusion models recently achieving state-of-the-art performance via iterative noise-to-data transformations.
  
 \textbf{Spectral-based Graph Generation}. Spectral information—eigenvalues and eigenvectors of a graph's adjacency or Laplacian matrix—encodes key structural properties such as connectivity and clustering. Leveraging this, SPECTRE \cite{martinkus2022} addresses the limitations of one-shot generative models by first generating dominant spectral components and then constructing graphs to match them. However, SPECTRE relies on three GANs with complex architectures and does not directly model eigenvector distributions; instead, it refines eigenvectors from a bank of learned matrices. Alternatively, GSDM \cite{luo2023fast} integrates graph spectrum into the diffusion framework by applying stochastic differential equations (SDEs) to eigenvalues, while sampling eigenvectors from ground truth data. GSDM avoids eigenvector diffusion due to challenges with applying SDEs on the Stiefel manifold. GGSD \citep{minellogenerating}, a contemporaneous work, further attempts to model the distribution of eigenvectors, but it disregards the manifold constraints. In contrast, we leverage geodesic flow matching to directly model eigenvector dynamics on the manifold.
  
  \textbf{Flow Matching on Manifolds}. Flow matching \cite{lipman2023flow, albergo2023building, liu2023flow} offers an efficient alternative to diffusion models by learning straighter probability paths, with strong results in image and video generation. Importantly, it can be extended to manifold-based data. However, existing manifold methods often suffer from costly simulations \cite{huang2022riemannian}, poor scalability \cite{ben2022matching}, or biased approximations \cite{de2022riemannian}. To overcome these limitations, \citet{chen2024flow} proposed Riemannian Flow Matching (RFM), a scalable framework using geodesic distances to define vector fields on manifolds. Building on RFM, we adapt this approach to graphs by constructing an eigenvector generator on the Stiefel manifold. A contemporaneous work \citep{cheng2024stiefel} also applies flow matching on the Stiefel manifold, focusing on molecular generation with fixed moments of inertia, rather than general graph generation.
  
  \section{Preliminaries}
  \subsection{Spectral Decomposition of Graphs}
  A graph $G$ is defined by a set of vertices $V=\{ v_1 ,\cdots,v_n \}$ and a set of edges $E=\{(x,y)|\,(x,y)\in V^2,(x,y) \in E\iff (y,x)\in E)\}$, indicating two vertices are connected or not. The adjacency matrix $A$ of the graph $G$ is an $n$-order square matrix defined as follows:
  \begin{align*}
  A_{ij}=\left\{
     \begin{aligned}
  1,&  & {\rm if}\, (v_i,v_j)\in E \\
  0,&  & {\rm if}\, (v_i,v_j)\notin E 
  \end{aligned}
  \right. 
  \end{align*} 
  The degree of a vertex is defined as the number of edges connected to it (or the sum of the weights of those edges), which is denoted by ${\rm deg}(v_i)$. The degree matrix is an $n$-order matrix defined as:
  \begin{align*}
  D_{ij}=\left\{
     \begin{aligned}
  &{\rm deg}(v_i),&  & {\rm if}\, i=j \\
  &0,&  & {\rm if}\, i\neq j 
  \end{aligned}
  \right. 
  \end{align*}
  The normalized Laplacian matrix is $L = I - D^{-\frac{1}{2}}AD^{-\frac{1}{2}}$. Its eigendecomposition yields a diagonal matrix of eigenvalues and an orthogonal matrix of eigenvectors, capturing key structural features of the graph. These spectral components provide intrinsic node coordinates and support metrics such as commute time \cite{fouss2007random} and biharmonic distances \cite{lipman2010biharmonic}. Thus, the Laplacian and its spectrum offer a natural, intrinsic graph representation.
  
  \subsection{Orthogonal Matrices and Stiefel Manifolds}
  \label{Orthgonal and Stiefel}
  The set of orthogonal matrices is defined as:
  \begin{align*}
  O(n) = {P \in M_n :, P^T P = I_n},
  \end{align*}
  which forms a Riemannian manifold of dimension $\frac{n(n-1)}{2}$, embedded in the Euclidean space $\mathbb{R}^{n \times n}$. Its exponential map coincides with the standard matrix exponential:
  \begin{align*}
  \exp(Q) = \sum_{i=0}^\infty \frac{Q^i}{i!},
  \end{align*}
  where $Q \in \mathbb{R}^{n \times n}$. As $n$ grows, the dimension of $O(n)$ increases quadratically, posing computational challenges. To mitigate this, we adopt the Stiefel manifold $\mathbb{V}_k(\mathbb{R}^{n})$, a lower-dimensional submanifold of $O(n)$, defined as:
  \begin{align*}
      \mathbb{V}_k(\mathbb{R}^{n})=\{U\in M_{n\times k}:\, U^TU=I_k  \}
  \end{align*}
  This manifold has a dimension of $n k-\frac{k(k+1)}{2}$, which grows linearly with $n$. The exponential map in $\mathbb{V}_k(\mathbb{R}^n)$ \citep{edelman1998geometry} is defined as:
  \begin{align*}
      &{\rm Exp}:T\mathbb{V}_k(\mathbb{R}^n)\to \mathbb{V}_k(\mathbb{R}^n)\\
      {\rm Exp}{(U,v)}&= (U,\,v)
  \exp
  \begin{pmatrix}
  A & -S(0) \\
  I & A
  \end{pmatrix}
  I_{2p,p} e^{-At}
  \end{align*}
  where $U\in\mathbb{V}_k(\mathbb{R}^n),\,v\in T_U\mathbb{V}_k(\mathbb{R}^n)$, $A=U^Tv$, $S(0)=v^Tv$, $I_{2p,p}=\begin{pmatrix}
  I_p \\
  0
  \end{pmatrix}_{2p\times p}$. Basic properties of Stiefel manifolds are introduced in Appendices \ref{Key_properties_stiefel}.
  
  \subsection{Flow Matching}
  \label{flow_matching}
  Given an vector field $u_t(\cdot)$ on a Riemannian manifold $\mathcal{M}$ that depends on time $t$, the flow mapping $\phi_t(\cdot)$ of $u_t(\cdot)$ is defined by the solution of the following ordinary differential equation (ODE):
  \begin{align*}
      &\phi_t(x_0)=x(t)\\
      {\rm where}\quad&\frac{dx}{dt}=u_t(x)
      ,\,x(0)=x_0
  \end{align*}
  Given an initial distribution, which is usually a white noise, on $\mathcal{M}$ with a density function $p(\cdot)$. Denote the push-forward measure $[\phi_t(\cdot)]_{*}p$ by $p_t$, the relationship between $p_t$
    and $u_t(x)$ is given by the continuity equation:
    \begin{align*}
        \frac{d}{dt}p_t=-{\rm div}(u_tp_t)
    \end{align*}
    
  The purpose of flow matching is to fit a designed vector field $u_t(x)$ by a neural network $V_{\theta}(t,x)$ that satisfies the condition: 
    \begin{align*}
        p_0(x)=p(x),\quad p_1(x)=q(x)
    \end{align*}
  where $q(x)$ is the density function of the desired measure. \citet{lipman2023flow} derived such a vector field $u_t(x)$ by firstly designing conditional vector fields $u_t(\cdot|x_1)$ for $x_1\in \mathcal{M}$ (or equivalently design a conditional flow mapping $\psi_t(\cdot|x_1)$, the flow mapping of $u_t(\cdot|x_1)$). Denoting the push-forward measure $\psi_t(\cdot|x_1)_{*}p$ by $p_t(x|x_1)$, the conditional flow mapping satisfies:
  \begin{align*}
      p_0(x|x_1)=p(x),\quad p_1(x|x_1)\approx \delta_{x_1}
  \end{align*}
  where $\delta_{x_1}$ is the Dirac distribution centered at $x_1$. Then, $u_t(\cdot)$ is determined by:
   \begin{align*}
      u_t(x)=\int_{\mathcal{M}} u_t(x|x_1)\frac{p_t(x|x_1)q(x_1)}{p_t(x)}\,d\omega\\
  \end{align*}
  where 
  \begin{align*}
      p_t(x)=\int_{\mathcal{M}} p_t(x|x_1)q(x_1)\,d\omega
  \end{align*}
  More importantly, using conditional vector fields or conditional flow mapping as supervision is sufficient for fitting $u_t(x)$. In practice, we optimize the following loss function to learn the vector field $V_\theta(t,x)$:
  \begin{align*}
      \mathcal{L}=\mathbb{E}_{t,q(x_1),p(x_0)}{\left\lVert V_{\theta}(t,x_t)-\dot{x}_t\right\rVert^2}
  \end{align*}
  where $\dot{x}_t=\frac{d}{dt}\psi_t(x_0|x_1)=u_t(x_t|x_1)$.
  
  \section{Method}
  SFMG comprises three flow-matching models (Figure~\ref{fig:pipeline}). The first two model the generation of eigenvalues $\Lambda = (\lambda_1,\dots,\lambda_k) \in \mathbb{R}^k$ and the corresponding eigenvector matrix $U \in \mathbb{V}_k(\mathbb{R}^n) \subset \mathbb{R}^{n \times k}$. Specifically, we learn vector fields that transport white noise to the data distribution:
  \begin{align*}
  \left\{
     \begin{aligned}
  &d\Lambda = V_{1,\theta}(\Lambda,t)\,dt\\
  &dU = V_{2,\theta}(U, \Lambda|_{t=1}, t)\,dt
  \end{aligned}
  \right.
  \end{align*}
  Here, $V_{2,\theta}$ is conditioned on the eigenvalues. While a full Laplacian requires all $n$ eigencomponents, we approximate it using the top $k$. The resulting matrix $L_0 = U\,{\rm diag}(\lambda_1,\dots,\lambda_k)\,U^T$ is not a valid Laplacian. To bridge this gap, we use the third flow matching model to transform it into a valid adjacency matrix and further matching the data distribution.:
  \begin{align*}
      dL = V_{3,\theta}(L,t)\,dt
  \end{align*}
  For graphs with node features, we jointly model the distributions of adjacency and feature matrices $F \in \mathbb{R}^{n \times m}$ via:
  \begin{align*}
  \left\{
     \begin{aligned}
  &dL = V_{3,\theta}(L,F,t)\,dt\\
  &dF = V_{3,\theta}(L,F,t)\,dt
  \end{aligned}
  \right.
  \end{align*}

  SFMG proceeds by first performing eigendecomposition of the normalized Laplacian:
  \begin{align*}
      L = P \cdot {\rm diag}(\lambda_1,\dots,\lambda_n) \cdot P^T
  \end{align*}
  This maps each graph to $O(n) \times \mathbb{R}^n$. For efficiency, we retain only the top $k$ eigenvectors and eigenvalues. Two flow matching models are then used to learn the distribution of the first $k$ eigenvalues in $\mathbb{R}^k$ and the corresponding eigenvectors in $\mathbb{V}_k(\mathbb{R}^n)$, conditioned on the eigenvalues. The third flow model reconstructs the adjacency matrix and learns node features from the generated eigenvalues $\lambda_1,\dots,\lambda_k$ and eigenvectors $U \in \mathbb{R}^{n \times k}$. Finally, symmetry and rounding are applied to ensure valid adjacency matrices (Appendix~\ref{sec:appendix-implem}). The following subsections describe the design and implementation of these models.

  \subsection{Conditional Flow Mapping Design}
  \label{Exp_Stiefel}
  As introduced in section \ref{flow_matching}, we only need to design conditional flow mapping to implement flow matching in $\mathbb{R}^k$ and $\mathbb{V}_k(\mathbb{R}^n)$.
  
  \textbf{Path in Euclidean space} 
  Euclidean space possesses a plane structure, allowing us to apply the simplest linear interpolation to conditional flow mapping:
  \begin{align*}
      \psi_t(x_0|x_1)=x_t=x_0+t(x_1-x_0)
  \end{align*}
  which corresponds to the flow of (conditionally) optimal transport:
  \begin{align*}
      u_t=\frac{d}{dt}\psi(x_0|x_1)=x_1-x_0
  \end{align*}
  
  \textbf{Path in Stiefel manifolds}.  
  The exponential map naturally defines geodesics in $\mathbb{V}_k(\mathbb{R}^n)$, which are introduced in section \ref{Orthgonal and Stiefel}. By ${\rm Exp}(U,\cdot)$ and its inverse ${\rm Log}(U,\cdot)$, we derive the geodesic interpolation between two points on $\mathbb{V}_k(\mathbb{R}^n)$. This interpolation is used to define conditional flow mapping in $\mathbb{V}_k(\mathbb{R}^n)$, as designed by \citet{chen2024flow}:
  \begin{align*}
      \psi_t(U_0|U_1)=x_t:={\rm Exp}{\big(U_0,t\cdot {\rm Log}(U_0,U_1)\big)}
  \end{align*}
  Its corresponding conditional vector field is (Appendices \ref{conditional_vec_field}):
  \begin{align*}
      u_t(\cdot|U_1)=\frac{d}{dt}\psi_t(U_0|U_1)={\rm Log}{(U_t,U_1)}\frac{\Vert{\rm Log}{(U_0,U_1)}\Vert}{\Vert{\rm Log}{(U_t,U_1)}\Vert}
  \end{align*}
  In practice, we compute the logarithm map of Stiefel manifolds using the algorithm proposed by \citet{zimmermann2022computing}.

  \subsection{Network Structure and Vector Field Learning}
  \label{Network Structure}
  Within the SFMG framework, we adopt task-specific architectures to learn vector fields $V_{\theta}$. For graph datasets, we use a ResNet-style network composed of residual blocks:
  \begin{align*}
      {\rm Block}_\theta(x)=\sigma\left(LN_1 \circ W_2 \circ \sigma\circ LN_2\circ W_1(x)+x\right)
  \end{align*}
  where $LN_1$ and $LN_2$ are layer normalization layers, $W_1$ and $W_2$ are linear maps, and $\sigma$ denotes ReLU. For molecular data, we follow the architecture used in GDSS.
  
  Since the tangent space of $\mathbb{R}^n$ is trivial, we directly use the network to learn vector fields in Euclidean space. In contrast, the tangent spaces of the Stiefel manifold $ \mathbb{V}_k(\mathbb{R}^n)$ are nontrivial, and the raw network output $Z$ may not lie on a valid tangent plane. To ensure validity, we project $Z$ onto the tangent space at point $Y$ using the projection operator: 
  \begin{align*}
      V_{2,\theta}(Y,t)=\pi_T(Z,Y)&=Y\frac{Y^TZ-Z^TY}{2}+(1-YY^T)Z
  \end{align*}
  as detailed in Appendix~\ref{Key_properties_stiefel}. This guarantees that the learned vector field respects the geometric structure of the Stiefel manifold.
  
  \subsection{Training and Generation}
  We learn conditional geodesic flows on $\mathbb{R}^k$, $\mathbb{V}_k(\mathbb{R}^n)$, and $\mathbb{R}^{n\times n}$ by minimizing the loss:
  \begin{align*}
      \resizebox{1.0\hsize}{!}{$
    L_{RCFM}=\mathbb{E}_{t,q(x_1),p(x_0)}\left\Vert V_{\theta}(x_t,t)-{\rm Log}(x_t,x_1)\frac{\Vert{\rm Log}(x_0,x_1)\Vert}{\Vert{\rm Log}(x_t,x_1)\Vert}\right\Vert^2
    $}
  \end{align*}
  where ${\rm Log}(x_0,x_1)=x_1-x_0$ in Euclidean space, and in $\mathbb{V}_k(\mathbb{R}^n)$, it denotes the inverse of the exponential map.

  \textbf{Training}.
  For the eigenvector generator, we condition the vector field on the first $k$ true eigenvalues. After training the first two flow models, we generate Laplacian matrices $L_0$, which are then transformed by a third model into adjacency matrices in $\mathbb{R}^{n\times n}$ and used to learn the feature matrix distribution. Training follows the method in \cite{lipman2023flow} (Algorithm \ref{train_eigenval},\ref{train_eigenvec},\ref{train_post}).

  \begin{algorithm}
\begin{algorithmic}
\caption{Training of flow matching for eigenvectors}
\label{train_eigenvec}
\STATE 1. Sample $t \sim U[0,1]$.  
Sample $U_1, \Lambda_1$ from data (first $k$ eigenvectors and eigenvalues).  
Sample $U_0$ from white noise on the Stiefel manifold.  

\STATE 2. Compute the flow matching loss:  
\[
\mathcal{L} = 
\left\Vert 
V_{\theta}(U_t, t \mid \Lambda_1) 
- {\rm Log}(U_t, U_1) \frac{\Vert {\rm Log}(U_0, U_1) \Vert}{\Vert {\rm Log}(U_t, U_1) \Vert}
\right\Vert^2
\]  
Perform a single optimization step to minimize $\mathcal{L}$.  

\STATE 3. Repeat Steps 1 and 2 until convergence.
\end{algorithmic}
\end{algorithm}

\begin{algorithm}
\begin{algorithmic}
\caption{Training of flow matching for eigenvalues}
\label{train_eigenval}

\STATE 1. Sample $t \sim U[0,1]$.  
Sample $\Lambda_1 = (\lambda_1, \dots, \lambda_k)$ from the data (first $k$ eigenvalues).  
Sample $\Lambda_0$ from $\mathcal{N}(0, I_k)$ (white noise).  

\STATE 2. Compute the flow matching loss:  
\[
\mathcal{L} = 
\left\Vert 
V_{\theta}(\Lambda_t, t) 
- \frac{\Vert \Lambda_1 - \Lambda_0 \Vert}{\Vert \Lambda_1 - \Lambda_t \Vert} (\Lambda_1 - \Lambda_t) 
\right\Vert^2
\]  
Perform a single optimization step to minimize $\mathcal{L}$.  

\STATE 3. Repeat Steps 1 and 2 until convergence.
\end{algorithmic}
\end{algorithm}

  \begin{algorithm}
  \begin{algorithmic}
  \caption{Training of flow matching for the postprocess}  \label{train_post}

  \STATE 1. Sample $t\sim U[0,1]$, $L_1$ from data of the adjacency matrix and $L_0=U{\rm diag}(\lambda_1,\cdots,\lambda_k)U^T$, where $U$ and $(\lambda_1,\cdots,\lambda_k)$ are generated by the eigenvector generator and eigenvalue generator respectively.\\
  
  \STATE 2. Perform optimization on the loss function 
  \[
  \left\Vert V_{\theta}(L_t,t)-\frac{\Vert L_1-L_0\Vert}{{\Vert L_1-L_t}\Vert}(L_1-L_t)\right\Vert^2,
  \]
  with a single step.
  
  Repeat Steps 1 and 2 until convergence.
  \end{algorithmic}
  \end{algorithm}

\begin{algorithm}[htp]
\begin{algorithmic}
\caption{Sampling from SFMG}
\label{sampling_}

\STATE 1. Sample $\Lambda_0 \sim \mathcal{N}(0, I_k)$ and apply the Euler method for $V_{1,\theta}$ to obtain $\Lambda_1$.

\STATE 2. Sample $U_0$ uniformly from $\mathbb{V}_k(\mathbb{R}^n)$ and apply the Euler method for $V_{2,\theta}(\cdot \mid \Lambda_1)$ to obtain $U_1$.

\STATE 3. Construct the initial adjacency matrix
\[
L_0 = U_1 \, \text{diag}(\Lambda_1) \, U_1^T,
\]
sample $F_0 \sim \mathcal{N}(0, I_{n \times m})$, and apply the Euler method for $V_{3,\theta}$ to obtain $L_1$ and $F_1$.

\STATE 4. Apply simple post-processing to $L_1$ and $F_1$ if needed.

\STATE \textbf{Return} adjacency matrix $L_1$ and feature matrix $F_1$.

\end{algorithmic}
\end{algorithm}
  
\textbf{Generation}.
  Sample generation proceeds in three stages: (1) sampling eigenvalues in $\mathbb{R}^k$; (2) generating eigenvectors in $\mathbb{V}_k(\mathbb{R}^n)$ conditioned on them; and (3) constructing adjacency matrices from the resulting Laplacian $L_0$ (Algorithm~\ref{sampling_}). These steps involve solving ODEs. We adopt the Euler method with step size $\epsilon$, combined with manifold exponential mapping:
  \begin{align*}
      x_{t+1} = {\rm Exp}(x_t, \epsilon \cdot V_t(x_t))
  \end{align*}
  where ${\rm Exp}(x,v) = x+v$ in $\mathbb{R}^n$, and is defined for $\mathbb{V}_k(\mathbb{R}^n)$ in Section~\ref{Exp_Stiefel}. Empirically, we find the generation quality robust to $\epsilon$, and set it to 0.01 in practice.

  % \begin{algorithm}[H]
  % \begin{algorithmic}
  % \label{sampling_}
  % \caption{Sampling}
  % \STATE \textbf{Step 1}. Sample $\Lambda_0$ from $\mathcal{N}(0,I_{k})$ and apply the Euler method for $V_{1,\theta}$ to derive $\Lambda_1$.
  % \STATE \textbf{Step 2}. Sample $U_0$ from uniform distribution of $\mathbb{V}_k(\mathbb{R}^n)$ and apply the Euler method for $V_{2,\theta}(\cdot|\Lambda_1)$ to derive $U_1$.
  
  % \STATE \textbf{Step 3}. Set $L_0$ by $U_1{\rm diag}(\Lambda_1)U_1^T$, sample $F_0$ from $\mathcal{N}(0,I_{n\times m})$ and apply the Euler method for $V_{3,\theta}$ to derive $L_1$ and $F_1$.
  
  % \STATE \textbf{Step 4} A simple Post-process.
  % \STATE \textbf{Return} adjacency matrix $L_1$ and feature matrix $F_1$.
  % \end{algorithmic}
  % \end{algorithm}

  \section{Experiments}
  \subsection{Datasets}
  To evaluate graph generation performance, we use a diverse set of datasets: six general graph benchmarks (Ego-Small, Community-Small, Planar, Enzymes, SBM, and Grid) and the molecular dataset QM9. These widely used datasets \cite{you2018graphrnn, niu2020, martinkus2022, vignac2023} cover a broad range of graph structures, enabling a comprehensive assessment of model generalization and adaptability (see Table~\ref{tab:genera_graph_dataset}). We follow standard practice by splitting each dataset into 80\% training and 20\% test sets.
\begin{table}[htbp]
  \caption{Statistics of graph datasets used in this study.}
  \label{tab:genera_graph_dataset}
  \vskip 0.15in
  \begin{center}
  \begin{tabular}{lcc}
      \toprule
      Name & \#Graphs & \#Nodes \\
      \midrule
      Ego-Small	        &200 & [4, 18]	\\ 
      Community-Small	    &100 & [12, 20]	\\
      Planar	            &200 & 64	     \\
      Enzymes             & 587 & [10, 125] \\
      SBM	                &200 & [44, 192]	 \\
      Grid	            &100 & [100, 400] \\
  \bottomrule
  \end{tabular}
  \end{center}
  \vskip -0.1in
  \end{table}

\subsection{Evaluation Metrics}

  % We assess graph generation quality by comparing statistical distributions of key properties between generated and test graphs. Specifically, we consider node degree (Deg.), clustering coefficient (Clus.), orbit count (Orbit), and normalized Laplacian eigenvalues (Spec.), using Maximum Mean Discrepancy (MMD) \cite{gretton2012} to quantify differences. To evaluate fit to the training distribution, we report the average MMD Ratio, with lower values indicating better alignment. Diversity is measured using Uniqueness (Uni.) and Novelty (Nov.). The number of generated samples matches the test set size. For the QM9 dataset, we follow GDSS and additionally compute Fréchet ChemNet Distance (FCD) \cite{preuer2018frechet}, NSPDK \cite{costa2010fast}, and MMD to assess both chemical and structural fidelity. Full metric definitions are in Appendix~\ref{sec:appendix-metric}.

We assess graph generation quality by comparing statistical distributions of key properties between generated and test graphs. Specifically, we consider:

\begin{itemize}
    \item \textbf{Node degree (Deg.)} and \textbf{clustering coefficient (Clus.)}: local graph statistics;
    \item \textbf{Orbit count (Orbit)}: counts of 4-node subgraph motifs capturing higher-order structures;
    \item \textbf{Normalized Laplacian eigenvalues (Spec.)}: global graph spectral information.
\end{itemize}

We quantify differences using \textbf{Maximum Mean Discrepancy (MMD)} \cite{gretton2012}, defined as
\[
\begin{aligned}
\mathrm{MMD}(\mathbb{S}_g, \mathbb{S}_r) &= 
\frac{1}{m^2} \sum_{i,j=1}^m k(\mathbf{x}_i^r, \mathbf{x}_j^r) 
+ \frac{1}{n^2} \sum_{i,j=1}^n k(\mathbf{x}_i^g, \mathbf{x}_j^g) \\
&\quad - \frac{2}{nm} \sum_{i=1}^n \sum_{j=1}^m k(\mathbf{x}_i^g, \mathbf{x}_j^r)
\end{aligned}
\]
where $k(\cdot,\cdot)$ is a kernel function; we use an RBF kernel with Earth Mover's Distance or total variation distance for efficiency.

To measure diversity, we report:

\begin{itemize}
    \item \textbf{Uniqueness (Uni.)}: fraction of generated graphs belonging to unique isomorphism classes;
    \item \textbf{Novelty (Nov.)}: fraction of generated graphs not present in the training set.
\end{itemize}

\textbf{Validity} is assessed for Planar and SBM graphs:  
a valid Planar graph is connected and planar; a valid SBM graph has 2–5 communities with 20–40 nodes per community, and estimated parameters match the original with probability $\ge 0.9$ \cite{martinkus2022}. Table~\ref{tab:vun} reports validity, uniqueness, and novelty results.

For QM9 molecules, we follow GDSS and additionally compute \textbf{Fréchet ChemNet Distance (FCD)} \cite{preuer2018frechet}, NSPDK \cite{costa2010fast}, and MMD to assess chemical and structural fidelity.

\subsection{Selection of $k$ for the Stiefel Manifold}
We adopt the configuration proposed in SPECTRE \cite{martinkus2022} to select the hyperparameter $k$ for the Stiefel manifold $\mathbb{V}_k(\mathbb{R}^n)$, aiming to ensure both improved performance and methodological fairness. 
The choice of $k$ is guided by three considerations: 

(i) \emph{Computational cost}. Large $k$ increases memory and training load, whereas using $k<n$ alleviates overhead since the complexity of $SO(n)$ grows quadratically with $n$, while that of the Stiefel manifold grows linearly; 

(ii) \emph{Learning capacity}. Empirical results show that optimization directly on $SO(n)$ often performs poorly, while the Stiefel manifold provides a more suitable geometry for high-dimensional features; and 

(iii) \emph{data characteristics}. Spectral theory indicates that small eigenvalues encode global structure (e.g., connectivity, clustering), so a modest $k$ is often sufficient, though larger graphs may require larger values. 

In practice, $k$ can be initialized following prior work (e.g., spectral decomposition with GANs to learn the first $k$ eigenvectors) and refined with simple search strategies such as binary search when necessary.
 \begin{table*}[htbp]
      \caption{Performance evaluation of SFMG and the competing methods on the Ego-Small and Community-Small datasets, respectively.Metrics include node degree (Deg.), clustering coefficient (Clus.), orbit counts (Orbit), and Laplacian spectra (Spec.), all measured by MMD. “Ratio” is the average MMD ratio (lower is better), and “Uniq. \& Nov.” measure generation diversity.}
      \label{tab:mmd-1}
      \centering
      \resizebox{\linewidth}{!}{
          \begin{tabular}{lcccc>{\columncolor{gray!15}}c>{\columncolor{gray!15}}ccccc>{\columncolor{gray!15}}c>{\columncolor{gray!15}}c}
              \toprule
              & \multicolumn{6}{c}{Ego-Small} & \multicolumn{6}{c}{Community-Small} \\ 
              \cmidrule(lr){2-7}
              \cmidrule(lr){8-13}
              Methods & {Deg.\,$\downarrow$} & {Clus.\,$\downarrow$} & {Orbit\, $\downarrow$} & {Spec.\,$\downarrow$} & Ratio\,$\downarrow$ & {Uniq. \& Nov.} & {Deg.\,$\downarrow$} & {Clus.\,$\downarrow$} & {Orbit\, $\downarrow$} & {Spec.\,$\downarrow$} & {Ratio\,$\downarrow$} & {Uniq. \& Nov.}  \\
              \midrule
              Training/Test & 0.0008        & 0.0234         & 0.0052         & 0.0066         & 1.0 & 32.5 & 0.0006        & 0.038          & 0.0025         & 0.0179         & 1.0 & 55.0\\
              \midrule
      GraphRNN       & 0.0137        & 0.0795         & 0.0401         & 0.0120          & 7.5    & 27.4 & 0.0116        & 0.3476         & 0.039          & 0.1163         & 12.6     & 96.8     \\ 
      GRAN           & 0.0085        & 0.0370         & 0.0180          & 0.0092         & 4.2       & 25.0  & 0.0042        & 0.1408         & 0.0345         & 0.0413         & 6.7     & 90.0 \\ 
      SPECTRE        & 0.0056        & 0.0609         & 0.0044         & 0.0118         & 3.0       & 57.5  & 0.0030         & 0.1799         & 0.0178         & 0.0615         & 5.0      &100.0  \\ 
      EDP-GNN        & 0.0092        & 0.0893         & 0.0128         & 0.0174         & 5.1      & 47.5  & 0.0073        & 0.1607         & 0.0769         & 0.0628         & 12.7    & 100.0 \\ 
      GDSS           & 0.0037        & 0.0318         & 0.0162         & 0.0110          & 2.7      & 15.0    & 0.0067        & 0.1613         & \textbf{0.0068}         & 0.0482         & 5.2     & 100.0 \\ 
      DiGress        & 0.0032        & \textbf{0.0205}         & 0.0124         & 0.0060          & 2.0      & 20.0   & 0.0029        & \textbf{0.1349}         & 0.0153         & 0.0331         & 4.1    & 90.0 \\ 
              \midrule
              Noise FM & 0.0024	& 0.0346 & \textbf{0.0028} & 0.0063 & 1.5 & 27.5 & \textbf{0.0021}	&0.1439	&0.0729	&0.0452	&9.7 & 100.0 \\
      \textbf{SFMG}  & \textbf{0.0015}	& 0.0319 &	0.0067	& \textbf{0.0048} &	\textbf{1.3} & 37.5 & 0.0022 & 0.1396 & 0.0141 & \textbf{0.0280} & \textbf{3.6} & 95.0\\
      \bottomrule
          \end{tabular}
      }
  \end{table*}
  
  \begin{table*}[htbp]
      \caption{Performance evaluation of SFMG and the competing methods on the Planar and Enzymes datasets, respectively. Metrics follow Table 1. Lower MMD/Ratio values indicate closer alignment, while higher Uniqueness and Novelty reflect greater diversity.}
      \label{tab:mmd-2}
      \centering
      \resizebox{\linewidth}{!}{
          \begin{tabular}{lcccc>{\columncolor{gray!15}}c>{\columncolor{gray!15}}ccccc>{\columncolor{gray!15}}c>{\columncolor{gray!15}}c}
              \toprule
              & \multicolumn{6}{c}{Planar} & \multicolumn{6}{c}{Enzymes} \\ 
              \cmidrule(lr){2-7}
              \cmidrule(lr){8-13}
              Methods & {Deg.\,$\downarrow$} & {Clus.\,$\downarrow$} & {Orbit\, $\downarrow$} & {Spec.\,$\downarrow$} & Ratio\,$\downarrow$ & {Uniq. \& Nov.} & {Deg.\,$\downarrow$} & {Clus.\,$\downarrow$} & {Orbit\, $\downarrow$} & {Spec.\,$\downarrow$} & {Ratio\,$\downarrow$} & {Uniq. \& Nov.}  \\
              \midrule
       Training/Test & 0.0002        & 0.031          & 0.0005         & 0.0052         & 1.0  & 100.0 & 0.0019        & 0.0136         & 0.0074         & 0.0100           & 1.0  & 100.0 \\
      \midrule
      GraphRNN       & 0.0049        & 0.2779         & 1.2543         & 0.0459         & 637.7     & 100.0   & 0.0110         & 0.0219          & 0.0222          & 0.0084         & 2.8  & 100.0            \\
      GRAN           & 0.0007        & \textbf{0.0426}         & \textbf{0.0009}         & 0.0075         & 2.0       & 2.5  & 0.0344        & \textbf{0.0168}         & 0.0230          & 0.0176         & 6.1     & 98.3    \\
      SPECTRE        & 0.0005        & 0.0785         & 0.0012         & 0.0112         & 2.4      & 100.0  & 0.2865        & 0.5381         & 0.0951         & 0.0176         & 51.2    & 100.0      \\
      EDP-GNN        & 0.0097        & 0.3583         & 1.6544         & 0.0899         & 846.5     & 100.0  & 0.0293       & 0.0962     & 0.1043    & 0.0423        & 10.2      & 100.0    \\
      GDSS           & 0.0455        & 0.2808         & 0.5314         & 0.0541         & 327.4     & 100.0   & 0.0097         & 0.0410          & 0.0269          & 0.0211             & 3.5     &  100.0   \\
      DiGress        & 0.0006        & 0.0563         & 0.0098         & \textbf{0.0062 }        & 6.4      & 100.0    & 0.0140         & 0.0741         & \textbf{0.0045}         & 0.0177         & 3.8    & 100.0     \\
      \midrule
      Noise FM & 0.1428 &	0.2596	& 1.1034 & 0.0320	& 733.8 & 100.0 & 0.1363	& 0.0310 & 	0.7089	& 0.0260 &	43.1 & 100.0 \\
      \textbf{SFMG}  & \textbf{0.0003} & 0.0613 & 0.0011 & 0.0084 & \textbf{1.9} & 72.5 & \textbf{0.0064} & 0.0352	& 0.0282 & \textbf{0.0065} & \textbf{2.6} & 54.7 \\
      \bottomrule
          \end{tabular}
      }
  \end{table*}

\subsection{Baselines}
To evaluate the effectiveness of SFMG, we compare it with several baseline methods that are representative of state-of-the-art graph generation methods. Specifically, GraphRNN \cite{you2018graphrnn} and GRAN \cite{liao2019} are two autoregressive models, while SPECTRE \cite{martinkus2022} is a one-shot GAN-based model. For diffusion-based models, EDP-GNN \cite{niu2020} and GDSS \cite{jo2022score} are continuous ones, and DiGress \cite{vignac2023} is a discrete one. 

\par We also consider models specifically designed for molecular generation like MolFlow \cite{zang2020molflow}, GraphAF \cite{shi2020graphaf}, and GraphDF \cite{luo2021graphdf} as baseline models.

\par GSDM \cite{luo2023fast}, another graph spectral diffusion model, was initially considered. However, we found that despite its fast generation speed, GSDM only learns the eigenvalues of the graph, which results in low diversity in the generated graphs. This limitation significantly reduces its ability to capture the full range of graph topologies, and as a result, we excluded GSDM from the direct baseline comparisons.

\subsection{Implementation Details}
For our proposed model, SFMG, we first preprocess the graphs for consistency. In a given dataset, we identify the maximum number of nodes and pad the adjacency matrix of each graph with zeros to match this maximum size (i.e., by adding isolated nodes).

\par Our spectral generation process is divided into two parts: eigenvalue generation and conditional eigenvector generation. We use an MLP with residual connections to learn the $\mathbb{R}^n$ vector field for eigenvalues. For eigenvectors, we use a separate MLP to learn the Stiefel Manifold vector field, conditioned on the generated eigenvalues. The specific network architectures for the eigenvalue generator, eigenvector generator, and post-processing network are detailed in Tables~\ref{tab:eigval_detail}, \ref{tab:eigvec_detail}, and \ref{tab:postprocess_detail}, respectively. Once the eigenvalues $\tilde{\lambda}$ and eigenvectors $\tilde{U}$ are generated, we reconstruct the graph's Laplacian matrix as $\tilde{L} = \tilde{U} \text{diag}(\tilde{\lambda}) \tilde{U}^T$.

\par For the QM9 dataset, each molecule is transformed into a weighted graph with node features \( X \in \{0, 1\}^{N \times F} \) and an adjacency matrix \( A \in \{0, 1, 2, 3\}^{N \times N} \). The matrix \( A \) encodes bond types. We learn the distributions of the eigenvalues and eigenvectors from the Laplacian matrix. In the post-processing stage, we use flow matching to directly learn the mapping from the spectrum to the weighted graph. This is followed by a quantization operation to obtain discrete bond types: values less than 0.5 are set to 0 (no bond), values in [0.5, 1.5) are set to 1 (single), values in [1.5, 2.5) are set to 2 (double), and values $\ge$ 2.5 are set to 3 (triple). For unweighted topological graphs, we use a simpler thresholding rule: values less than 0.5 are set to 0, and values greater than or equal to 0.5 are set to 1.
  
 % \subsection{Baselines and Implementation Details}
 %  To evaluate the effectiveness of SFMG, we compare it with several baseline methods that are representative of state-of-the-art graph generation methods. Specifically, GraphRNN \cite{you2018graphrnn} and GRAN \cite{liao2019} are two autoregressive models, while SPECTRE \cite{martinkus2022} is a one-shot GAN-based model. For diffusion-based models, EDP-GNN \cite{niu2020} and GDSS \cite{jo2022score} are continuous ones, and DiGress \cite{vignac2023} is a discrete one. GSDM \cite{luo2023fast} is a graph spectral diffusion model that runs diffusion SDEs on the graph spectrum space. However, we found that despite its fast generation speed, GSDM only learns the eigenvalues of the graph, which results in low diversity in the generated graphs (Appendix \ref{GSDM v.s. SFMG}). This limitation significantly reduces its ability to capture the full range of graph topologies. As a result, we decided to exclude GSDM from the baseline comparisons. Besides, we consider models specifically designed for molecular generation like MolFlow \cite{zang2020molflow}, GraphAF \cite{shi2020graphaf}, and GraphDF \cite{luo2021graphdf} as baseline models. For SFMG, we employ neural network structures as depicted in Section \ref{Network Structure}, and please refer to Appendix~\ref{sec:appendix-implem} for further implementation details.
  
  \begin{table*}[htbp]
      \caption{Performance evaluation of SFMG and the competing methods on the SBM and Grid datasets, respectively. Metrics are consistent with Tables 1–2. SFMG achieves strong spectral fidelity and competitive diversity compared with baselines.}
      \vspace{0.125in}
      \label{tab:mmd-3}
      \centering
      \resizebox{\linewidth}{!}{
          \begin{tabular}{lcccc>{\columncolor{gray!15}}c>{\columncolor{gray!15}}ccccc>{\columncolor{gray!15}}c}
              \toprule
              & \multicolumn{6}{c}{SBM} & \multicolumn{5}{c}{Grid} \\ 
              \cmidrule(lr){2-7}
              \cmidrule(lr){8-12}
              Methods & {Deg.\,$\downarrow$} & {Clus.\,$\downarrow$} & {Orbit\, $\downarrow$} & {Spec.\,$\downarrow$} & Ratio\,$\downarrow$ & {Uniq. \& Nov.} & {Deg.\,$\downarrow$} & {Clus.\,$\downarrow$} & {Orbit\, $\downarrow$} & {Spec.\,$\downarrow$} & {Uniq. \& Nov.}  \\
              \midrule
       Training/Test & 0.0008        & 0.0332         & 0.0255         & 0.0063         & 1.0  & 100.0 & 0           & 0           & 0           & 0.0112 & 10.0 \\
      \midrule
          GraphRNN       & 0.0055        & 0.0584         & 0.0785         & 0.0065         & 3.2     & 100.0   & 0.0041      & 0.0002      & 0.0012      & 0.0146        & 95.9     \\
          GRAN           & 0.0113        & 0.0553         & 0.0540          & 0.0054         & 4.7     & 100.0   & \textbf{0.0008}      & 0.0037      & 0.0015      & 0.0162        & 90.0     \\
          SPECTRE        & 0.0015        & 0.0521         & 0.0412         & 0.0056         & 1.5    & 100.0     & 0.3650       & \textbf{0.0001}      & 0.6795          & 0.0516             & 95.0       \\
          EDP-GNN        & 0.0015        & 0.0590          & 0.0677         & 0.0064         & 1.8     & 100.0     & 0.2581      & 0.6466      & 0.9736      & 0.0471         & 100.0    \\
          GDSS           & 0.1702        & 0.0911         & 0.2268         & 0.0629         & 58.6    & 100.0     & 0.1205      & 0.1406      & 0.0835      & 0.0440          & 100.0   \\
          DiGress        & 0.0013        & \textbf{0.0498}         & 0.0434         & 0.0400           & 2.8     & 100.0    & 0.0053      & 0.0044      & 0.0214      & \textbf{0.0101}          & 100.0    \\
      \midrule
      Noise FM &\textbf{0.0005}	&0.0554	&0.0566	&0.0068	&1.4 & 100.0 & 0.3364	& 0.8177	& 1.1574	& 0.0380  & 100.0 \\
      \textbf{SFMG}  & 0.0008	& 0.0511 & \textbf{0.0386} & \textbf{0.0035} & \textbf{1.2} & 67.5 & 0.0020 &	0.0004 &	\textbf{0.0007} &	0.0132  & 75.0 \\
      \bottomrule
          \end{tabular}
      }
      \vspace{-0.1in}
  \end{table*}

\subsection{Results}
\subsubsection{Graph Benchmarks}
\begin{figure*}[htbp]
  \vskip 0.2in
  \begin{center}
  \centerline{\includegraphics[width=\textwidth]{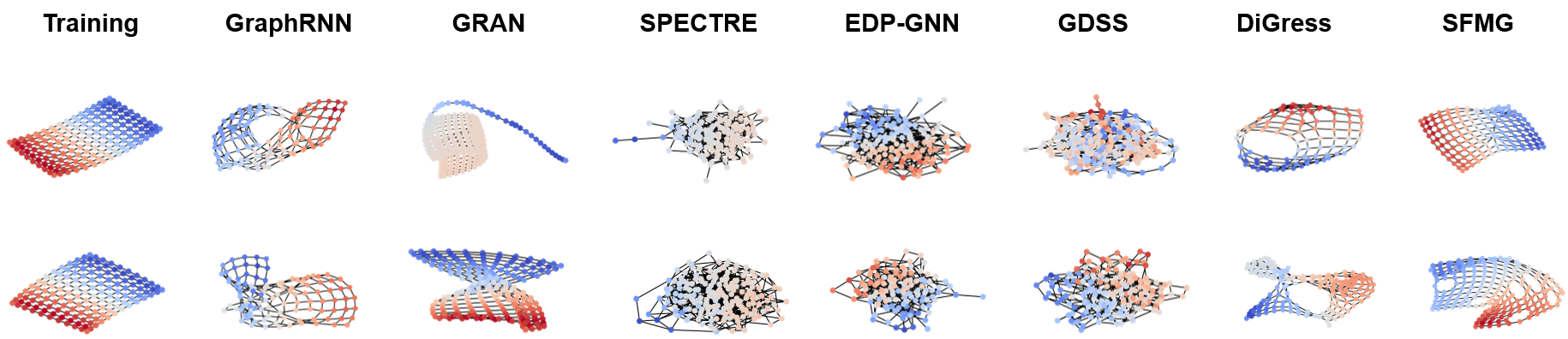}}
  \caption{Compared to other approaches, SFMG produces graphs with clearer structures and topologies that most closely resemble the ground truth.}
  \label{fig:samples}
  \end{center}
  \vskip -0.2in
  \end{figure*}

We evaluate SFMG against baselines on diverse graph datasets (Tables~\ref{tab:mmd-1}–\ref{tab:mmd-3}). 
SFMG consistently achieves the highest Ratio scores, demonstrating its ability to generate structurally and spectrally faithful graphs. It also outperforms baselines on spectral MMD in most cases, indicating effective 
spectrum modeling. Notably, SFMG surpasses diffusion-based models such as EDPGNN and GDSS on large graphs (e.g., Grid), and performs comparably to the state-of-the-art autoregressive model GraphRNN. In terms of diversity, SFMG remains competitive, and qualitative samples 
(Figure~\ref{fig:samples}) further support its performance.

While most baselines achieve near 100\% uniqueness and novelty (Uniq. \& Nov.) across datasets, SFMG ranges from 55\% to 95\%. However, due to the discrete nature of graphs, perfect uniqueness may indicate excessive deviation from the training distribution, potentially compromising structural coherence. In contrast, SFMG’s moderate novelty reflects a desirable balance between diversity and fidelity to underlying data patterns.

To further evaluate the quality of generated graphs, we report Validity, Uniqueness, and Novelty (VUN) metrics specifically for Planar and SBM datasets (Table~\ref{tab:vun}). These datasets have well-defined structural constraints: a Planar graph is valid if it is connected and planar, and an SBM graph is valid if its community structure matches the generative parameters. Other datasets lack strict constraints, so VUN metrics are less informative. Table~\ref{tab:vun} shows that SFMG achieves higher validity than baselines while maintaining strong diversity, highlighting its ability to generate both structurally correct and diverse graphs.

\begin{table}[htbp]
    \caption{Validity, Uniqueness, and Novelty (V.U.N) results of different methods on Planar and SBM graphs.}
    \vspace{0.125in}
    \label{tab:vun}
    \centering
    \setlength{\tabcolsep}{6pt} % standard spacing
    % Planar block
    \begin{tabular}{lcccc}
        \toprule
        \multicolumn{5}{c}{Planar} \\
        \cmidrule(lr){1-5}
        Method & Validity & Uniq. & Nov. & V.U.N \\
        \midrule
        GraphRNN & 0.0 & 100.0 & 100.0 & 0.0 \\
        SPECTRE  & 25.0 & 100.0 & 100.0 & 25.0 \\
        GDSS     & 0.0 & 100.0 & 100.0 & 0.0 \\
        SFMG     & 42.5 & 97.5  & 72.5  & 17.5 \\
        \bottomrule
    \end{tabular}
    
    \vspace{0.2cm} % small vertical gap between blocks
    
    % SBM block
    \begin{tabular}{lcccc}
        \toprule
        \multicolumn{5}{c}{SBM} \\
        \cmidrule(lr){1-5}
        Method & Validity & Uniq. & Nov. & V.U.N \\
        \midrule
        GraphRNN & 5.0 & 100.0 & 100.0 & 5.0 \\
        SPECTRE  & 52.5 & 100.0 & 100.0 & 52.5 \\
        GDSS     & 0.0 & 100.0 & 100.0 & 0.0 \\
        SFMG     & 75.0 & 97.5  & 67.5  & 35.0 \\
        \bottomrule
    \end{tabular}
\end{table}

  \begin{table}[htbp]
      \caption{Generation results on the QM9 dataset. Results are taken from original papers or GDSS. Relative results for SPECTRE and DiGress are not available in the original publication and are therefore indicated by `-'.}
      \centering
      \begin{tabular}{lcccc}
      \toprule
      Method & Val. (\%)$\uparrow$ & NSPDK$\downarrow$ & FCD$\downarrow$ & Time (s)$\downarrow$ \\
      \midrule
          GraphAF & 67.00 & 0.020 & 5.268 & 2.52$e^{3}$ \\
          GraphDF & 82.67 & 0.063 & 10.816 & 5.35$e^{4}$ \\
          MoFlow & 91.36 & 0.017 & 4.467 & 4.60 \\
          EDP-GNN & 47.52 & 0.005 & \textbf{2.680} & 4.40$e^{3}$ \\
          SPECTRE & 87.30 & - & - & -\\
          GDSS & 95.72 & \textbf{0.003} & 2.900 & 1.14$e^{2}$ \\
          DiGress & \textbf{99.00} & - & - & - \\
      \midrule
      \textbf{SFMG} & 85.67 & 0.005 & 2.750 & \textbf{3.51}\\
      \bottomrule
      \end{tabular}
      \label{tab:qm9_res}
  \end{table}

\subsubsection{Molecule Generation}
On the molecular dataset QM9 (Table~\ref{tab:qm9_res}), 
SFMG delivers competitive results, matching or surpassing baselines in validation accuracy, NSPDK, and FCD. 
It also excels in generation speed, requiring only 3.51 seconds, making it both effective and efficient.

  % For molecule generation, as shown in Table \ref{tab:qm9_res}, SFMG delivers competitive results, matching or surpassing baseline methods in terms of validation accuracy, NSPDK, and FCD. Additionally, it excels in generation speed, taking only 3.51 seconds, making it a highly efficient and effective method.

\subsubsection{Spectral Fidelity}
  Following \citet{martinkus2022}, we assess the quality of generated eigenvalues and eigenvectors by computing the direct MMD for the eigenvalues and the Wavelet MMD for the eigenvectors (Table ~\ref{tab:eigen}). 
  For eigenvalues, we also list the random results (i.e., randomly generate $k$ values that lie in the range [0,2]) as a reference, and we can see that SFMG (`SFMG' column) performs much better than the random generation and SPECTRE method. For eigenvectors, we consider two settings to evaluate the effectiveness of our eigenvector generator: generate eigenvectors conditioned on real eigenvalues (`Real' column in Table ~\ref{tab:eigen}) or generated eigenvalues (SFMG). The results in Table ~\ref{tab:eigen} show that SFMG achieves similar performance with Real and much better than SPECTRE. Overall, our method can effectively generate a graph spectrum.
  
  % \begin{table}[htbp]
  %     \caption{MMD ratios (vs training set MMD) for generated $k$ eigenvalues (direct MMD) and eigenvectors (Wavelet MMD).}
  %    \vspace{0.125in}
  %     \label{tab:eigen}
  %     \centering
  %         \resizebox{\linewidth}{!}{%
  %         \begin{tabular}{lcc>{\columncolor{gray!15}}ccc>{\columncolor{gray!15}}cc}
  %             \toprule
  %             & & \multicolumn{3}{c}{Eigenvalue} & \multicolumn{3}{c}{Eigenvector} \\ 
  %             \cmidrule(lr){3-5}
  %             \cmidrule(lr){6-8}
  %             Dataset & $k$ & Random & SFMG & SPECTRE & Real & SFMG &  SPECTRE \\
  %             \midrule
  %         Ego-Small   & 2  & 43.44  & 3.43 & 34.01 & 1.92  & 2.05  & 247.76 \\
  %         Community-Small  & 2  & 6.38  & 1.64  & 3.94 & 2.50 & 3.29 & 30.17  \\
  %         Planar   & 2  & 30.80 & 1.54 & 17.60 & 0.98 & 1.08 & 206.15 \\
  %         Enzymes    & 2  & 11.76 & 1.09  & 2.71 & 0.93  & 1.25   & 278.64    \\
  %         SBM      & 4  & 20.00 & 1.73 & 9.39 & 2.33 & 3.12  & 19.46   \\
  %         Grid     & 16 & 28.02 & 2.54 & 3.83 & 2.15  & 3.22  & 200.33    \\
  %         \bottomrule
  %         \end{tabular}%
  %         }
  %     \vspace{-0.1in}
  % \end{table}

\begin{table}[htbp]
\caption{MMD ratios (vs training set MMD) for generated $k$ eigenvalues (direct MMD) and eigenvectors (Wavelet MMD).}
\label{tab:eigen}
\centering
\begin{tabular}{lccccc}
\toprule
\multirow{2}{*}{Dataset} & \multirow{2}{*}{$k$} & \multicolumn{3}{c}{Eigenvalues (direct MMD)} \\ 
\cmidrule(lr){3-5}
 & & Random & SFMG & SPECTRE \\
\midrule
Ego-Small          & 2  & 43.44 & 3.43 & 34.01 \\
Community-Small    & 2  & 6.38  & 1.64 & 3.94  \\
Planar             & 2  & 30.80 & 1.54 & 17.60 \\
Enzymes            & 2  & 11.76 & 1.09 & 2.71  \\
SBM                & 4  & 20.00 & 1.73 & 9.39  \\
Grid               & 16 & 28.02 & 2.54 & 3.83  \\
\midrule
\multicolumn{2}{c}{} & \multicolumn{3}{c}{Eigenvectors (Wavelet MMD)} \\
\midrule
Ego-Small          & 2 & 1.92  & 2.05  & 247.76 \\
Community-Small    & 2 & 2.50  & 3.29  & 30.17  \\
Planar             & 2 & 0.98  & 1.08  & 206.15 \\
Enzymes            & 2 & 0.93  & 1.25  & 278.64 \\
SBM                & 4 & 2.33  & 3.12  & 19.46  \\
Grid               & 16 & 2.15  & 3.22  & 200.33 \\
\bottomrule
\end{tabular}
\end{table}

\subsubsection{Efficiency Analysis}
We compare the average time required to generate ten graphs using 
SPECTRE (GAN-based), GDSS (diffusion-based), and SFMG (flow matching-based). 
As reported in Table~\ref{tab:sample-time}, SPECTRE is the fastest (0.014–0.302s) due to the inherent efficiency of GAN sampling. 
GDSS is the slowest (18.244–102.870s), reflecting the iterative computational overhead of diffusion-based methods. 
SFMG, our flow matching-based approach, strikes a practical balance: its generation times (0.140–0.912s) are over $100\times$ faster than GDSS, yet slightly slower than SPECTRE. 
Importantly, despite this modest increase in time compared to GANs, SFMG delivers superior graph generation quality, demonstrating that flow matching provides both high fidelity and scalable performance.

 % \begin{table}[htbp]
 %      \caption{Time costs of three different methods for generating (sampling) ten graphs.}
 %      \vspace{0.125in}
 %      \label{tab:sample-time}
 %      \resizebox{\linewidth}{!}{
 %      \centering
 %          \begin{tabular}{lcccccc}
 %              \toprule
 %              Methods & Ego-Small & Community-Small & Planar & 
 %              Enzymes & SBM & Grid \\
 %              \midrule
 %              SPECTRE & 0.014 ± 0.000 & 0.014 ± 0.000 & 0.015 ± 0.000 & 0.031 ± 0.000 & 0.078 ± 0.000 & 0.302 ± 0.000 \\
 %              GDSS & 18.244 ± 0.383 & 39.024 ± 0.532 & 37.544 ± 0.429 & 54.355 ± 0.498 & 63.973 ± 0.418 & 102.870 ± 0.232 \\ 
 %              SFMG & 0.140 ± 0.002 & 0.168 ± 0.017 & 0.158 ± 0.006 & 0.190 ± 0.014 & 0.223 ± 0.022 & 0.912 ± 0.030 \\
 %      \bottomrule
 %          \end{tabular}
 %          }
 %      \vspace{-0.1in}
 %  \end{table}

\begin{table*}[htbp]
\caption{Time costs (seconds) of three different methods for generating ten graphs across six datasets.}
\label{tab:sample-time}
\centering
\resizebox{\textwidth}{!}{%
\begin{tabular}{lcccccc}
\toprule
Methods & Ego-Small & Community-Small & Planar & Enzymes & SBM & Grid \\
\midrule
SPECTRE & 0.014 ± 0.000 & 0.014 ± 0.000 & 0.015 ± 0.000 & 0.031 ± 0.000 & 0.078 ± 0.000 & 0.302 ± 0.000 \\
GDSS    & 18.244 ± 0.383 & 39.024 ± 0.532 & 37.544 ± 0.429 & 54.355 ± 0.498 & 63.973 ± 0.418 & 102.870 ± 0.232 \\
SFMG    & 0.140 ± 0.002 & 0.168 ± 0.017 & 0.158 ± 0.006 & 0.190 ± 0.014 & 0.223 ± 0.022 & 0.912 ± 0.030 \\
\bottomrule
\end{tabular}%
}
\end{table*}

  % The effectiveness and efficiency of SFMG are anticipated. SFMG directly utilizes flow matching, an advanced generative model, and fully leverages the structural information of the manifolds. These factors contribute to its effectiveness. Besides, SFMG is highly efficient, being more than 100 times faster than GDSS (Appendix \ref{sampling_efficiency}), a diffusion-based model, due to flow matching's ability to de-randomize the generation process. Although SFMG is approximately 10 times slower than SPECTRE, the time consumption remains acceptable.

 \subsubsection{Ablation studies}
  SFMG comprises two components: spectral generation and post-processing. To assess the contribution of the spectral generation module, we ablate it and instead apply flow matching directly from noise to the adjacency matrix distribution. This variant, denoted as Noise FM, is evaluated in Tables~\ref{tab:mmd-1}, \ref{tab:mmd-2}, and \ref{tab:mmd-3}. While Noise FM surpasses some baselines on some datasets, it consistently underperforms relative to SFMG, confirming the effectiveness of the spectral generation module in enhancing overall graph generation quality.

  \section{Conclusion}
  We introduced SFMG, a geometry-aware graph generative model that integrates spectral decomposition with Riemannian flow matching. By modeling eigenvectors on the Stiefel manifold and eigenvalues in Euclidean space, SFMG captures the spectral structure of graphs, while enabling efficient learning of graph distributions. Experiments on seven datasets show that SFMG achieves high-quality, scalable graph generation, outperforming diffusion models in speed, and matching the performance of leading autoregressive approaches. Our results highlight the potential of combining spectral methods with manifold-aware generative techniques for advancing structure-aware graph modeling.

  Although our method demonstrates strong performance on small to medium-sized graphs, it faces notable limitations when scaling to larger graphs with thousands of nodes. This is primarily due to the computational complexity of high-dimensional optimization on the Stiefel manifold and the intensive cost of solving manifold-based ordinary differential equations (ODEs). Although geodesic flow matching facilitates distribution learning on certain manifolds, its extension to high-dimensional regimes remains nontrivial. These challenges highlight the need for more expressive and scalable manifold-specific architectures.
  
\section*{Acknowledgments}
This work has been supported by the Strategic Priority Research Program of the Chinese Academy of Sciences [No. XDB0680101 to S.Z.], the CAS Project for Young Scientists in Basic Research [No. YSBR-034 to S.Z.], the National Natural Science Foundation of China [Nos. 32341013, 12326614], and the Robotic AI-Scientist Platform of the Chinese Academy of Sciences.

  \bibliographystyle{elsarticle-num-names}
  \bibliography{reference}
  \begin{IEEEbiography}[{\includegraphics[width=1in,height=1.25in,clip,keepaspectratio]{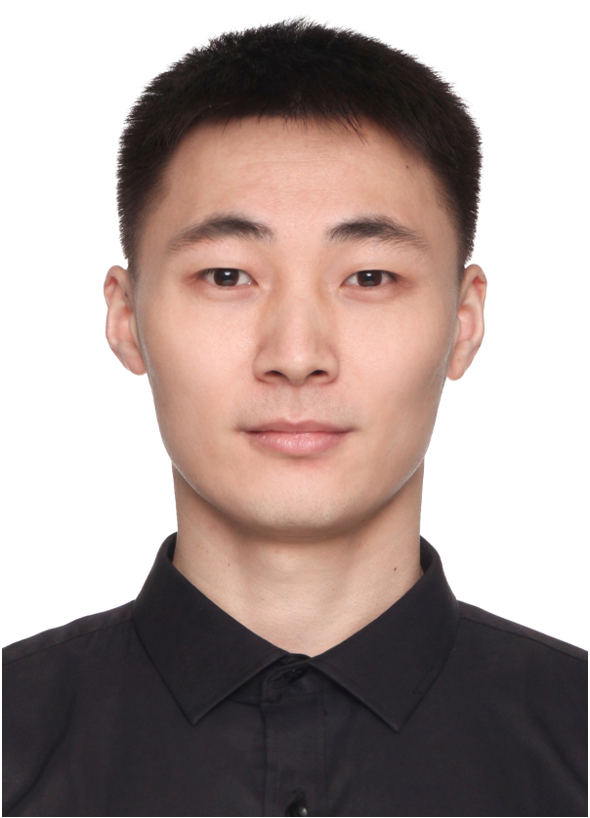}}]{Xikun Huang} has worked as a postdoc at the State Key Laboratory of Mathematical Sciences, Academy of Mathematics and Systems Science, Chinese Academy of Sciences. He received his PhD in Computer Science from the Academy of Mathematics and Systems Science, Chinese Academy of Sciences. He has moved to the Changping Laboratory as an Assistant Professor. His research interests include deep generative models, graph generation and artificial intelligence-driven drug design.
\end{IEEEbiography}

\begin{IEEEbiography}[{\includegraphics[width=1in,height=1.25in,clip,keepaspectratio]{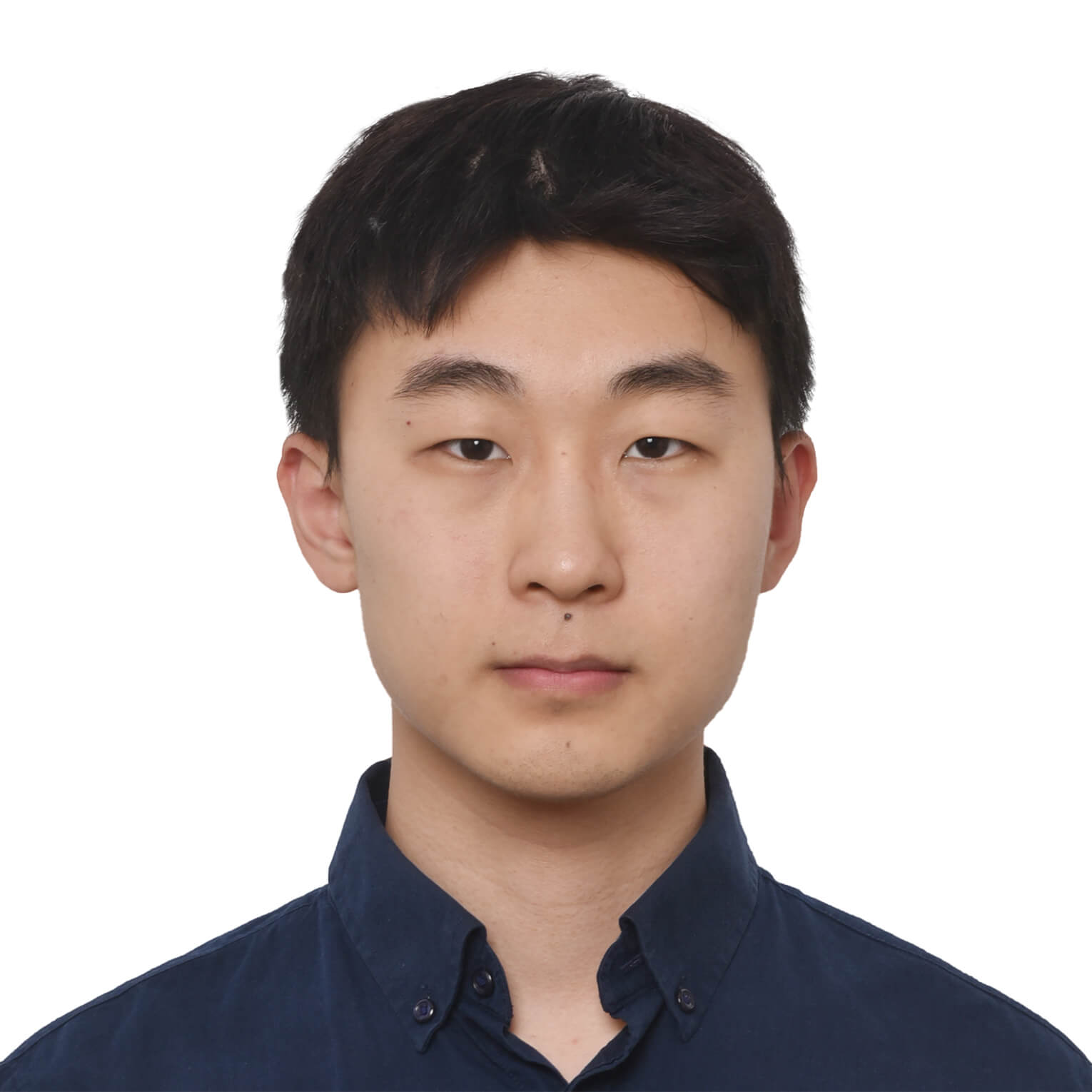}}]{Tianyu Ruan} is a PhD candidate in Operations Research and Cybernetics at the Academy of Mathematics and Systems Science, Chinese Academy of Sciences. His research interests include deep learning theory and geometry-inspired machine learning methods.
\end{IEEEbiography}

\begin{IEEEbiography}[{\includegraphics[width=1in,height=1.25in,clip,keepaspectratio]{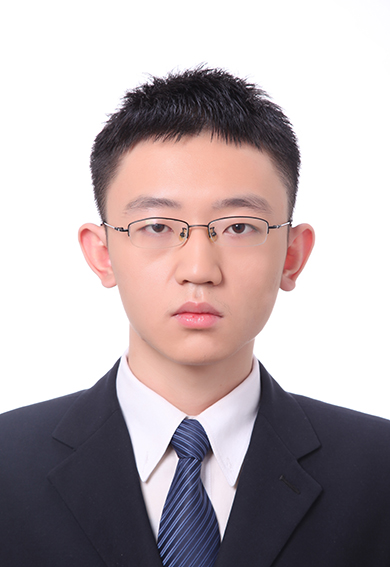}}]{Chihao Zhang} is an Assistant Professor with the Academy of Mathematics and Systems Science, Chinese Academy of Sciences. 
His research interests include machine learning, data mining, pattern recognition and bioinformatics.
\end{IEEEbiography}

\begin{IEEEbiography}[{\includegraphics[width=1in,height=1.25in,clip,keepaspectratio]{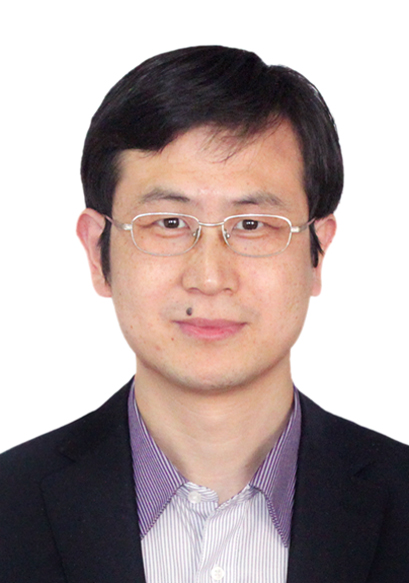}}]{Shihua Zhang} received the Ph.D. degree in applied mathematics and bioinformatics from the Academy of Mathematics and Systems Science, Chinese Academy of Sciences in 2008 with the highest honor. He joined the same institute as an Assistant Professor in 2008 and is currently Professor. His research interests are mainly in bioinformatics and computational biology, data mining, machine learning, and deep learning. He has won various awards and honors including the National High-level Talent Program Leading Talents (2022) and Top-notch Talents (2018), NSFC for Excellent Young Scholars (2014), Outstanding Young Scientist Program of CAS (2014), and Youth Science and Technology Award of China (2013). Now he serves as a Section Editor of PLOS Computational Biology.
\end{IEEEbiography}
  
  %%%%%%%%%%%%%%%%%%%%%%%%%%%%%%%%%%%%%%%%%%%%%%%%%%%%%%%%%%%%%%%%%%%%%%%%%%%%%%%
  %%%%%%%%%%%%%%%%%%%%%%%%%%%%%%%%%%%%%%%%%%%%%%%%%%%%%%%%%%%%%%%%%%%%%%%%%%%%%%%
  % APPENDIX
  %%%%%%%%%%%%%%%%%%%%%%%%%%%%%%%%%%%%%%%%%%%%%%%%%%%%%%%%%%%%%%%%%%%%%%%%%%%%%%%
  %%%%%%%%%%%%%%%%%%%%%%%%%%%%%%%%%%%%%%%%%%%%%%%%%%%%%%%%%%%%%%%%%%%%%%%%%%%%%%%
  \appendices

  \section{Key Properties of Stiefel Manifold $\mathbb{V}_k(\mathbb{R}^n)$}
  \label{Key_properties_stiefel}
  In this section, we introduce the fundamental properties of $\mathbb{V}_k(\mathbb{R}^n)=\{A\in M_{n\times k}:A^T\cdot A=I_{k\times k}\}$ as a Riemannian manifold (see \citet{edelman1998geometry} for more information). These properties provide insights into how to study vector fields on the Stiefel manifold and compute geodesics.
  
  \begin{Property}[Tangent space of $\mathbb{V}_k(\mathbb{R}^n)$]
  $T_Y\mathbb{V}_k(\mathbb{R}^n)=YA+(1-YY^T)B$, where $Y\in \mathbb{V}_k(\mathbb{R}^n)$, $A$ is a skew-symmetric matrix and $B$ is arbitrary $n\times k$ matrix.
  \end{Property}
  
  \textbf{Proof}: Suppose $\gamma(t)$ is a differentiable curve of $\mathbb{V}_k(\mathbb{R}^n)$ and $\gamma(0)=Y$. We have:
  \begin{align*}
      \gamma(t)^T\cdot\gamma(t)=I_k
  \end{align*}
  Take the derivative of both sides of the equation, we have:
  \begin{align*}
      \gamma(t)^T\cdot\dot{\gamma}(t)+\dot{\gamma}^T(t)\cdot\gamma(t)=0
  \end{align*}
  Let $t=0$, we have
  \begin{align*}
      \dot{\gamma}(t)^TY+Y^T\dot{\gamma}(t)=0
  \end{align*}
  As a result, $T_Y\mathbb{V}_k(\mathbb{R}^n)=\{C: C^TY+Y^TC\}=0$, since the dimension of $\{C: C^TY+Y^TC\}$ equals $nk-\frac{k(k+1)}{2}$. It is easy to check that \begin{align*}
      &\{YA,\,A \text{ is a skew-symmetric matrix} \}\\
      &\{(1-YY^T)B, B \text{ is arbitrary } n\times k \text{ matrix}\}
  \end{align*}
  are two subspaces of $T_Y\mathbb{V}_k(\mathbb{R}^n)$. It is clear that $\{(1-YY^T)B\}$ has $(n-k)k$ degrees of freedom, $\{YA\}$ has $\frac{k(k-1)}{2}$ degrees of freedom and they are orthogonal. Hence we complete the proof.
  
  $\hfill\square$

  \begin{Property}[Normal space of $\mathbb{V}_k(\mathbb{R}^n)$]
  $N_Y\mathbb{V}_k(\mathbb{R}^n)=\{YA,\,A\text{ is a symmetric }k\times k\text{ matrix}\}$, where $Y\in \mathbb{V}_k(\mathbb{R}^n)$.
  \end{Property}
  
  \textbf{Proof}: It is easy to check that 
  \begin{align*}
      \{YA,\,A\text{ is a symmetric }k\times k\text{ matrix}\}\perp T_Y\mathbb{V}_k(\mathbb{R}^n)
  \end{align*}
  Note that the dimension of $T_Y\mathbb{V}_k(\mathbb{R}^n)=\frac{k(k+1)}{2}$, we complete the proof.
  
  $\hfill\square$
  
  By these properties, it is easy to check that the projection of a matrix $Z\in M_{n\times k}$ to $N_Y\mathbb{V}_k(\mathbb{R}^n)$ and $T_Y\mathbb{V}_k(\mathbb{R}^n)$ is:
  \begin{align*}
      \pi_N(Z,Y)&=Y\frac{Y^TZ+Z^TY}{2}\\
      \pi_T(Z,Y)&=Y\frac{Y^TZ-Z^TY}{2}+(1-YY^T)Z
  \end{align*}
  
  \begin{Property}[Geodesics of $\mathbb{V}_k(\mathbb{R}^n)$]
  \begin{align*}
      &{\rm Exp}:T\mathbb{V}_k(\mathbb{R}^n)\to \mathbb{V}_k(\mathbb{R}^n)\\
      {\rm Exp}{(x,v)}&= (x,\,v)
  \exp
  \begin{pmatrix}
  A & -S(0) \\
  I & A
  \end{pmatrix}
  I_{2p,p} e^{-At}.
  \end{align*}
  where $x\in\mathbb{V}_k(\mathbb{R}^n),\,v\in T_x\mathbb{V}_k(\mathbb{R}^n)$, $A=x^Tv$, $I_{2p,p}=\begin{pmatrix}
  I_p \\
  0
  \end{pmatrix}_{2p\times p}$ and 
  \begin{align*}
      \exp{(Y)}=\sum_{i=0}^\infty \frac{Y^i}{i!}
  \end{align*}
  where $Y$ is a $n\times n$ matrix. By ${\rm Exp}(x,\cdot)$, we derive the geodesics of Stiefel manifold, and obtain interpolation between two reasonable points on $\mathbb{V}_k(\mathbb{R}^n)$.
  \end{Property}
  
  \textbf{Proof}: By definition, any geodesics $\gamma(t)$ of $\mathbb{V}_k(\mathbb{R}^n)$ satisfies:
  \begin{align*}
      \gamma(t)^T\gamma(t)=I_{k\times k}
  \end{align*}
  By taking derivatives twice, we have
  \begin{align*}
      \gamma(t)^T\ddot\gamma(t)+2\dot{\gamma(t)}^T\dot{\gamma(t)}+\ddot\gamma(t)^T\gamma(t)=0
  \end{align*}
  Since $\gamma(t)$ is geodesic, $\ddot\gamma(t)$ must be in normal space of $\gamma(t)$:
  \begin{align*}
      \ddot\gamma(t)+\gamma(t)S=0
  \end{align*}
  Combine this two equation we have:
  \begin{equation}
      \ddot\gamma(t) + \gamma(\dot{\gamma}^T\dot{\gamma})=0
      \label{eq_geo}
  \end{equation}
  Define
  \begin{align*}
      C(t)=\gamma^T\gamma,\quad A(t)=\gamma^T\dot\gamma\quad S=\dot\gamma^T\dot\gamma
  \end{align*}
  By Equation \ref{eq_geo}, we have:
  \begin{align*}
      &\dot{C}=A^T+A\\
      &\dot{A}=-CS+S\\
      &\dot{S}=[A,S]
  \end{align*}
  where $[A,S]=AS-SA$. Since $A$ is skew-symmetric, we have:
  \begin{align*}
      &C=I\\
      &A=A(0)\\
      &S=e^{At}S(0)e^{-At}
  \end{align*}
  By these equations, we know that:
  \begin{align*}
      \frac{d}{dt}(Ye^{At},\dot{Y}e^{At})=(Ye^{At},\dot{Y}e^{At})\begin{pmatrix}
  A & -S(0) \\
  I & A
  \end{pmatrix}
  \end{align*}
  By integration, we complete the proof.
  
  $\hfill\square$

  \section{Geodesics and the conditional vector field}
  \label{conditional_vec_field}
  To avoid introducing complex concepts, here we provide an intuitive understanding to the property used in section \ref{Exp_Stiefel}. Define the geodesic interpolation as:
  \begin{align*}
      \psi_t(U_0|U_1)=x_t:={\rm Exp}{\big(U_0,t\cdot {\rm Log}(U_0,U_1)\big)}
  \end{align*}
  Its corresponding conditional vector field is given by:
  \begin{align*}
      u_t(\cdot|U_1)=\frac{d}{dt}\psi_t(U_0|U_1)={\rm Log}{(U_t,U_1)}\frac{\Vert{\rm Log}{(U_0,U_1)}\Vert}{\Vert{\rm Log}{(U_t,U_1)}\Vert}
  \end{align*}
  \textbf{Intuitive Explanation:} When everything is well-defined, ${\rm Exp}{\big(U_0,t\cdot {\rm Log}(U_0,U_1)\big)}$ represents the $t$-quantile on the shortest path connecting $U_0$ and $U_1$ (the point on the shortest path where the distance from $U_0$ to this point, divided by the total distance from $U_0$ to $U_1$, equals $t$). A particle moving along this path exhibits the following property: Starting from $U_0$ and moving along the shortest path towards $U_1$, if the particle first travels a distance $d_0$ to reach $x_{t_0}$, and then moves along the shortest path from $x_{t_0}$ to $U_1$ for a distance $d_1$, the resulting position $x_{t_0+\Delta t}$ will be the same as if it had traveled the total distance $d_0+d_1$ directly along the path. Therefore, we have:
\begin{align*}
\psi_{t_0+\Delta t}(U_0|U_1) 
   &= {\rm Exp}\!\left(U_0,(t_0+\Delta t)\cdot 
        {\rm Log}(U_0,U_1)\right) \nonumber \\
   &= {\rm Exp}\!\left(x_{t_0}, d_1\cdot 
        \frac{{\rm Log}(U_{t_0},U_1)}
             {\lVert{\rm Log}(U_{t_0},U_1)\rVert}\right).
\end{align*}
  where $\Vert \Delta t\cdot {\rm Log}(U_0,U_1)\big) \Vert = d_1$. Hence,
  % \begin{align*}
  %     \frac{d}{dt}\psi_t(U_0|U_1)&=\lim_{\Delta t\to 0}\frac{(\psi_{t+\Delta t}-\psi_t)(U_0|U_1)}{\Delta t}\\
  %     &=\lim_{\Delta t\to 0}\frac{{\rm Exp}{\big(x_{t},d_1\cdot \frac{{\rm Log}{(U_{t},U_1)}}{\Vert{\rm Log}{(U_{t},U_1)}\Vert} \big)}-x_t}{\Delta t}\\
  %     &=\lim_{\Delta t\to 0}\frac{{\rm Exp}{\big(x_{t},\Delta t\cdot\Vert{\rm Log}(U_0,U_1)\big) \Vert\frac{{\rm Log}{(U_{t},U_1)}}{\Vert{\rm Log}{(U_{t},U_1)}\Vert} \big)}-x_t}{\Delta t}\\
  %     &={\rm Log}{(U_t,U_1)}\frac{\Vert{\rm Log}{(U_0,U_1)}\Vert}{\Vert{\rm Log}{(U_t,U_1)}\Vert}
  % \end{align*}
\begin{align*}
    v_t &:= {\rm Log}(U_t, U_1), \quad v_0 := {\rm Log}(U_0, U_1) \\
    \frac{d}{dt}\psi_t(U_0|U_1) 
        &= \lim_{\Delta t\to 0} \frac{
            {\rm Exp}\big(x_t, d_1 \cdot \frac{v_t}{\|v_t\|}\big) - x_t
          }{\Delta t} \\
        &= \lim_{\Delta t\to 0} \frac{
            {\rm Exp}\big(x_t, \Delta t \cdot \|v_0\| \cdot \frac{v_t}{\|v_t\|}\big) - x_t
          }{\Delta t} \\
        &= {\rm Log}(U_t, U_1) \frac{\|v_0\|}{\|v_t\|}
\end{align*}
  % \newpage
  % \section{Training and Generation Algorithm}
  % \label{implementation_train_and_generation}
  % \subsection{Training}

  % \subsection{Sampling}

  \section{Experimental Details}

  \begin{table*}[htbp]
  \centering
  \caption{Architecture details of eigenvalue generator.}
  \resizebox{\linewidth}{!}{%
  \begin{tabular}{c|l|cccccc}
  \toprule
  & Hyperparameter & Ego-small & Community-small & Planar & Enzymes & SBM & Grid \\ 
  \midrule
  \multirow{4}{*}{ResidualMLP} & Hidden dimension & 128 & 32 & 32 & 32 & 32 & 128 \\
  & Number of residual blocks & 2 & 2 & 2 & 2 & 2 & 4 \\
  & Number of trainable parameters & 66.95K & 4.45K & 4.45K & 4.45K & 4.58K & 136.59K \\
  & Number of epochs & 1000000 & 100000 & 10000 & 100000 & 100000 & 200000 \\
  & Batchsize & 32 & 32 & 32 & 32 & 32 & 32 \\
  \midrule
  \multirow{2}{*}{Optimization} & Optimizer & AdamW & AdamW & AdamW & AdamW & AdamW & AdamW \\
  & Learning rate & $5.0 \times 10^{-4}$ & $1.0 \times 10^{-4}$ & $1.0 \times 10^{-4}$ & $1.0 \times 10^{-4}$ & $1.0 \times 10^{-4}$ & $1.0 \times 10^{-4}$ \\
  \bottomrule
  \end{tabular}
  }
  % \vspace{-0.2cm}
  \label{tab:eigval_detail}
  \end{table*}
  
  \begin{table*}[htbp]
  \centering
  \caption{Architecture details of eigenvector generator.}
  \resizebox{\linewidth}{!}{
  \begin{tabular}{c|l|ccccccc}
  \toprule
  & Hyperparameter & Ego-small & Community-small & Planar & Enzymes & SBM & Grid \\ 
  \midrule
  \multirow{4}{*}{ResidualMLP} & Hidden dimension & 512 & 512 & 256 & 1024 & 256 & 512 \\
  & Number of residual blocks & 4 & 4 & 4 & 2 & 4 & 4 \\
  & Number of trainable parameters & 2.14M & 2.15M & 0.60M & 4.72M & 0.92M & 8.04M \\
  & Number of epochs & 100000 & 20000 & 20000 & 20000 & 10000 & 20000 \\
  & Batchsize & 32 & 32 & 32 & 32 & 32 & 40 \\
  \midrule
  \multirow{2}{*}{Optimization} & Optimizer & AdamW & AdamW & AdamW & AdamW & AdamW & AdamW  \\
  & Learning rate & $1.0 \times 10^{-4}$ & $1.0 \times 10^{-4}$ & $5.0 \times 10^{-4}$ & $1.0 \times 10^{-4}$ & $1.0 \times 10^{-4}$ & $1.0 \times 10^{-3}$ \\
  \bottomrule
  \end{tabular}
  }
  % \vspace{-0.2cm}
  \label{tab:eigvec_detail}
  \end{table*}
  
  \begin{table*}[htbp]
  \centering
  \caption{Architecture details of postprocess.}
  \resizebox{\linewidth}{!}{
  \begin{tabular}{c|l|ccccccc}
  \toprule
  & Hyperparameter & Ego-small & Community-small & Planar & Enzymes & SBM & Grid \\ 
  \midrule
  \multirow{4}{*}{ResidualMLP} & Hidden dimension & 64 & 128 & 512 & 512 & 512 & 512 \\
  & Number of residual blocks & 2 & 4 & 4 & 8 & 4 & 4 \\
  & Number of trainable parameters & 0.05M & 0.23M & 6.31M & 20.23M & 37.95M & 136.74M \\
  & Number of epochs & 6000 & 6000 & 8000 & 30000 & 2000 & 10000 \\
  & Batchsize & 80 & 20 & 8 & 128 & 20 & 32 \\
  \midrule
  \multirow{2}{*}{Optimization} & Optimizer & AdamW & AdamW & AdamW & AdamW & AdamW & AdamW  \\
  & Learning rate & $1.0 \times 10^{-4}$ & $1.0 \times 10^{-4}$ & $1.0 \times 10^{-3}$ & $1.0 \times 10^{-4}$ & $5.0 \times 10^{-4}$ & $1.0 \times 10^{-4}$ \\
  \bottomrule
  \end{tabular}
  }
  % \vspace{-0.2cm}
  \label{tab:postprocess_detail}
  \end{table*}

  \subsection{Exclusion of GSDM as a Baseline.}
  \label{GSDM v.s. SFMG}
  
  \begin{table}[htbp]
  \centering
  \caption{Comparison of SFMG and GSDM on Community-small Dataset}
  \label{GSDM1}
  \begin{tabular}{lcccc}
  \toprule
  \textbf{Method} & \textbf{Degree $\downarrow$} & \textbf{Clustering $\downarrow$} & \textbf{Orbit $\downarrow$} & \textbf{Uniq+Nov} \\
  \midrule
  GSDM  & 0.0306 & 0.0331 & 0.0033 & \textbf{50.0} \\
  SFMG  & 0.0022 & 0.1396 & 0.0141 & \textbf{95.0} \\
  \bottomrule
  \end{tabular}
  \end{table}
  
  \vspace{1em}
  
  \begin{table}[htbp]
  \centering
  \caption{Comparison of SFMG and GSDM on Grid Dataset}
  \label{GSDM2}
  \begin{tabular}{lcccc}
  \toprule
  \textbf{Method} & \textbf{Degree $\downarrow$} & \textbf{Clustering $\downarrow$} & \textbf{Orbit $\downarrow$} & \textbf{Uniq+Nov} \\
  \midrule
  GSDM  & 0.0001 & 0.0000 & 0.0001 & \textbf{50.0} \\
  SFMG  & 0.0020 & 0.0004 & 0.0007 & \textbf{75.0} \\
  \bottomrule
  \end{tabular}
  \end{table}
  As shown in Table \ref{GSDM1} and \ref{GSDM2}, although GSDM reports competitive graph generation scores, we ultimately excluded it as a baseline due to two primary limitations. First, it demonstrates low generation novelty: both its uniqueness and novelty metrics are notably low (approximately 50), suggesting a high degree of output redundancy and potential overfitting. Second, GSDM models only the eigenvalues while disregarding the eigenvectors, resulting in a significant loss of structural information and an incomplete spectral representation. This design choice undermines its ability to capture the full complexity of graph structures. In contrast, our method (SFMG) jointly models eigenvalues and eigenvectors, enabling richer spectral encoding and improved generalization. Supporting experimental results are presented in the subsequent sections.
  
  % \section{Sampling Efficiency}
  % \label{sampling_efficiency}
  % We compare the average time required to generate ten graphs using three different methods: SPECTRE (GAN-based), GDSS (diffusion-based), and SFMG (our flow matching-based model). As shown in Table ~\ref{tab:sample-time}, SPECTRE achieves the fastest generation across all datasets, with times ranging from 0.014 to 0.302 seconds, leveraging the efficiency of GANs in sampling. GDSS has the slowest generation times, ranging from 18.244 to 102.870 seconds, as it involves a computationally expensive iterative sampling process typical of diffusion models. SFMG achieves a balance between speed and potential quality, with times ranging from 0.140 to 0.912 seconds, being significantly faster than GDSS and only slightly slower than SPECTRE. This demonstrates SFMG's efficiency and potential for scalable graph generation.
  
 % \newpage
  
  \section{Visualization of Generated Graphs}
\begin{figure*}[htbp]
  \vskip 0.2in
  \begin{center}
  \centerline{\includegraphics[width=\textwidth]{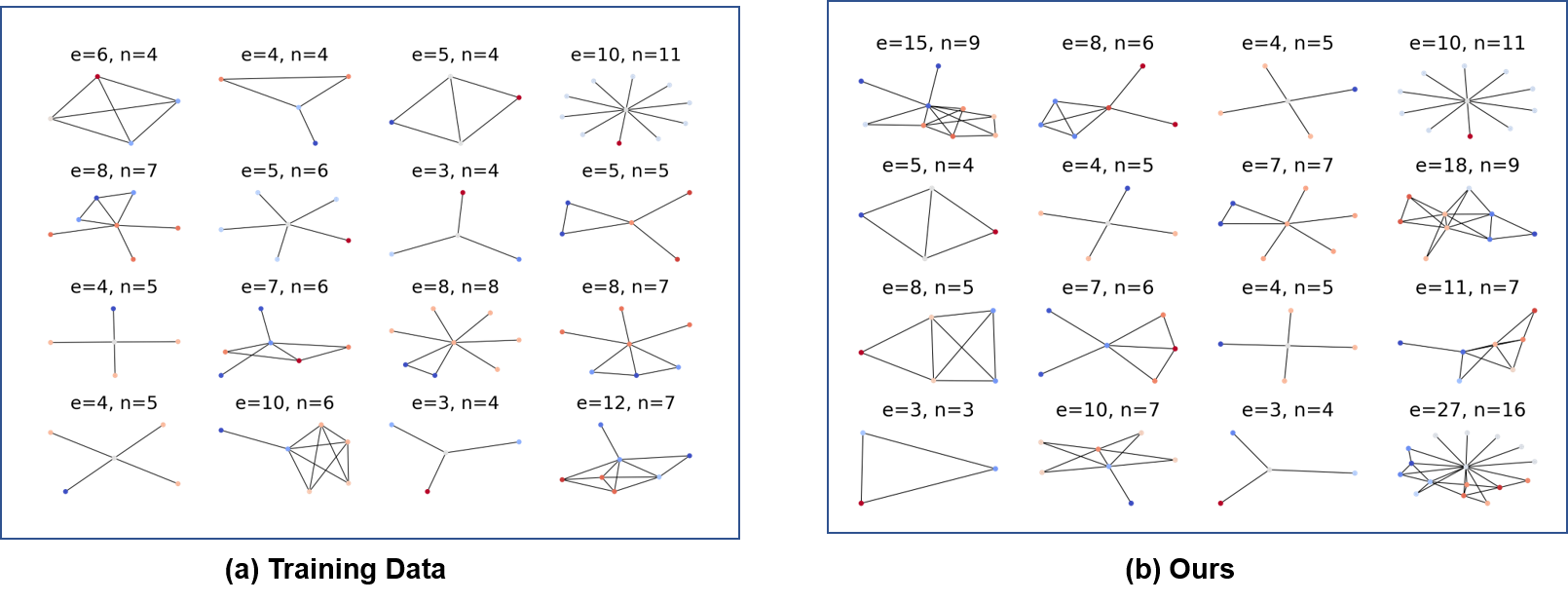}}
  \caption{Randomly selected Ego-Small graphs from the training and generated ones by SFMG.}
  \label{fig:ego-samples}
  \end{center}
  \vskip -0.2in
  \end{figure*}

\begin{figure*}[htbp]
  \vskip 0.2in
  \begin{center}
  \centerline{\includegraphics[width=\textwidth]{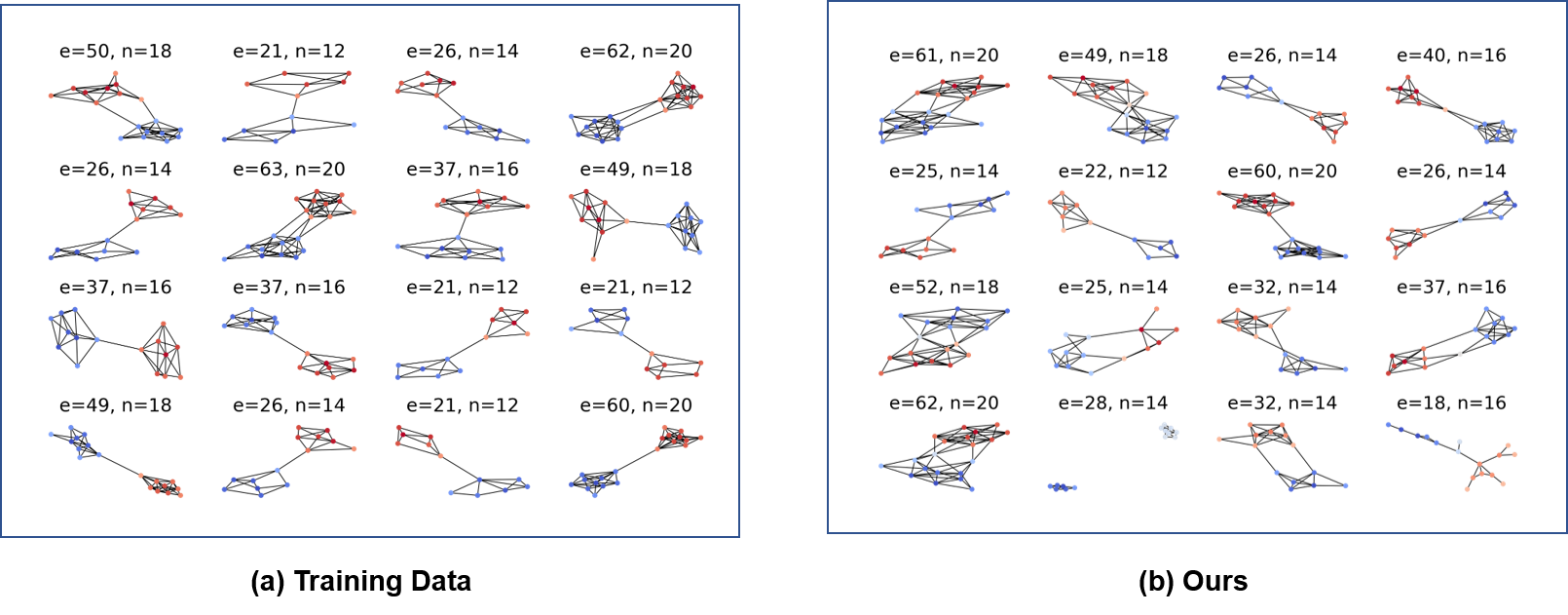}}
  \caption{Randomly selected Community-Small graphs from the training and generated ones by SFMG.}
  \label{fig:comm-samples}
  \end{center}
  \vskip -0.2in
  \end{figure*}

\begin{figure*}[htbp]
  \vskip 0.2in
  \begin{center}
  \centerline{\includegraphics[width=\textwidth]{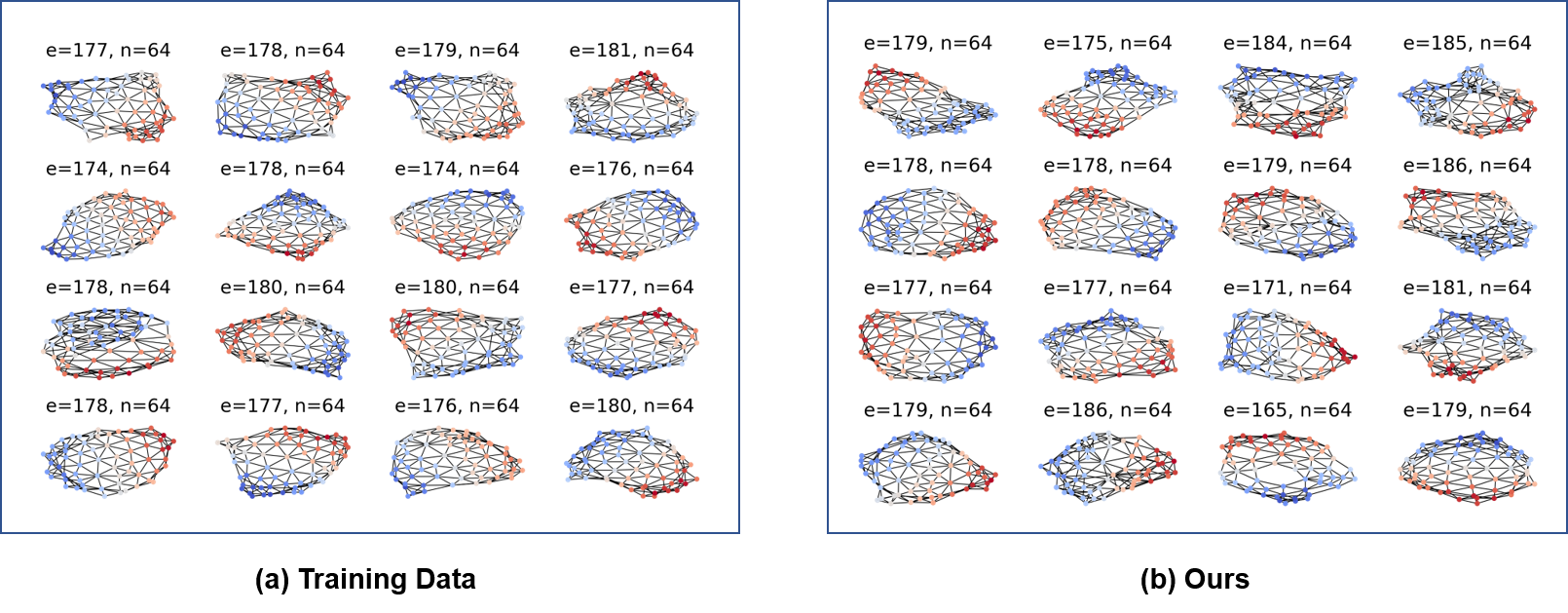}}
  \caption{Randomly selected Planar graphs from the training and generated ones by SFMG.}
  \label{fig:planar-samples}
  \end{center}
  \vskip -0.2in
  \end{figure*}

\begin{figure*}[htbp]
  \vskip 0.2in
  \begin{center}
  \centerline{\includegraphics[width=\textwidth]{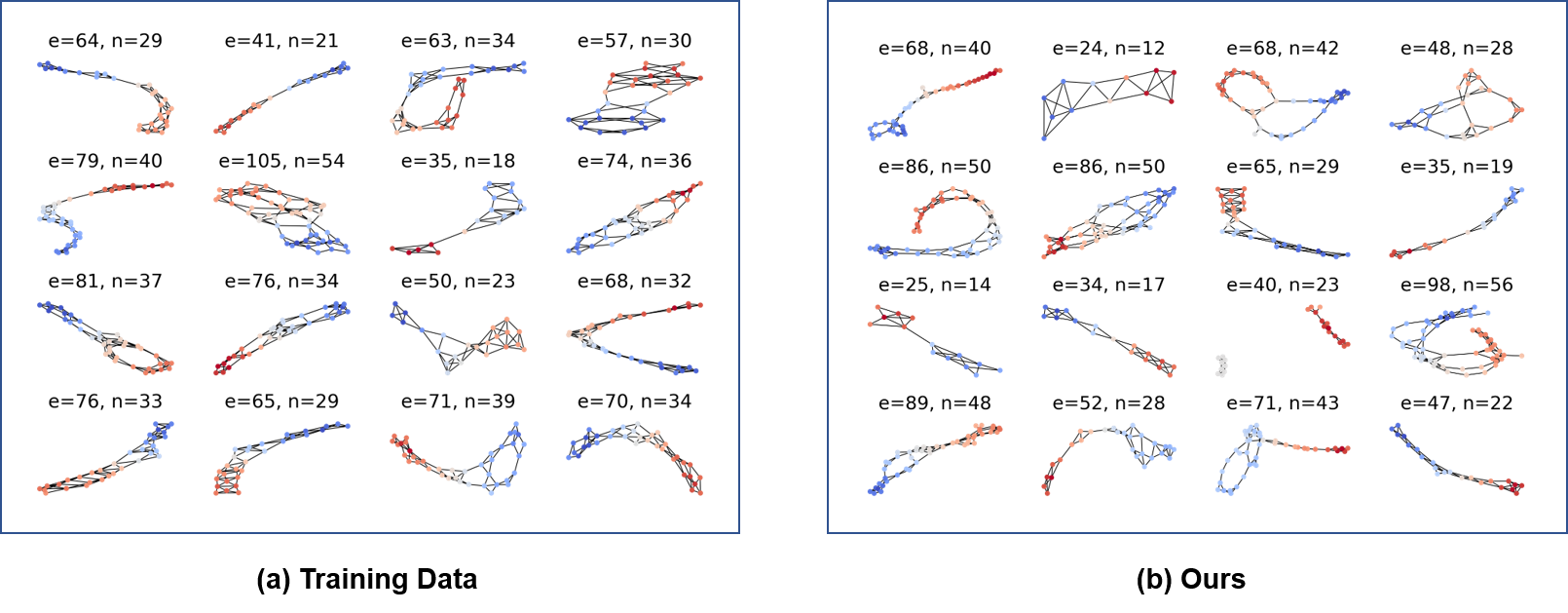}}
  \caption{Randomly selected Enzymes graphs from the training and generated ones by SFMG.}
  \label{fig:enzymes-samples}
  \end{center}
  \vskip -0.2in
  \end{figure*}

\begin{figure*}[ht]
  \vskip 0.2in
  \begin{center}
  \centerline{\includegraphics[width=\textwidth]{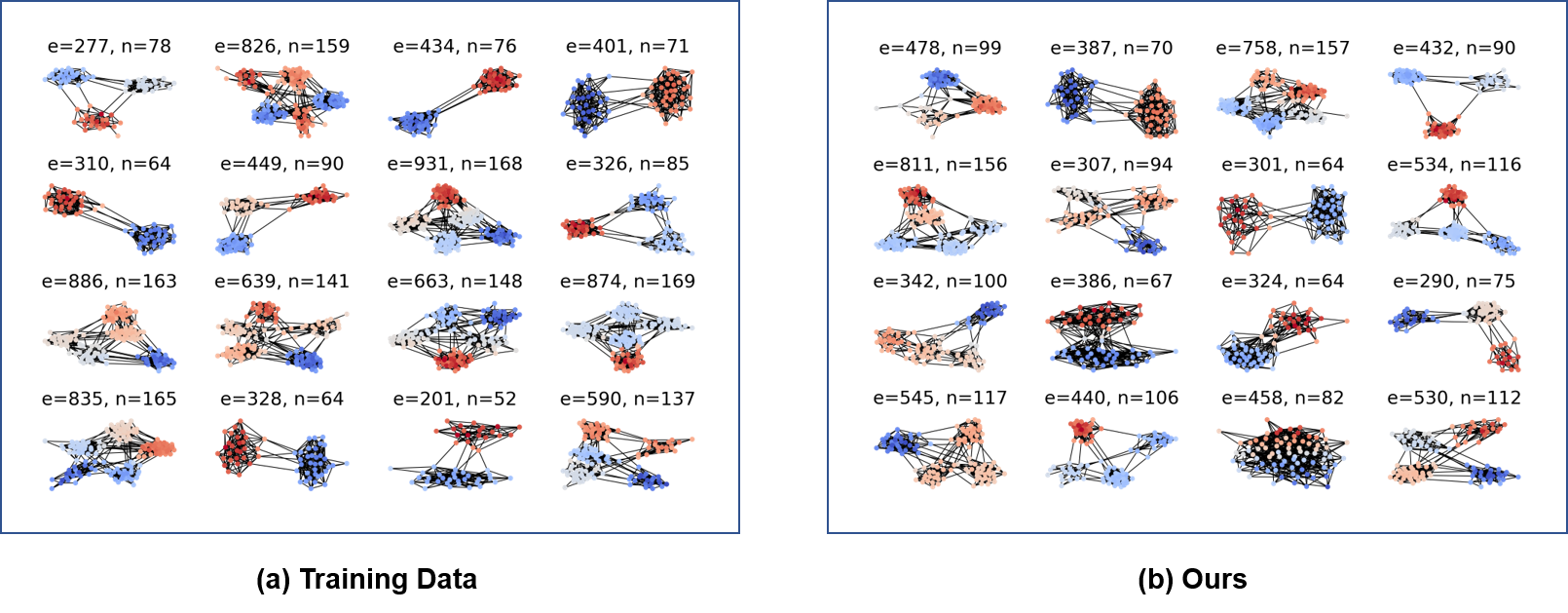}}
  \caption{Randomly selected SBM graphs from the training and generated ones by SFMG.}
  \label{fig:sbm-samples}
  \end{center}
  \vskip -0.2in
  \end{figure*}

\begin{figure*}[ht]
  \vskip 0.2in
  \begin{center}
  \centerline{\includegraphics[width=\textwidth]{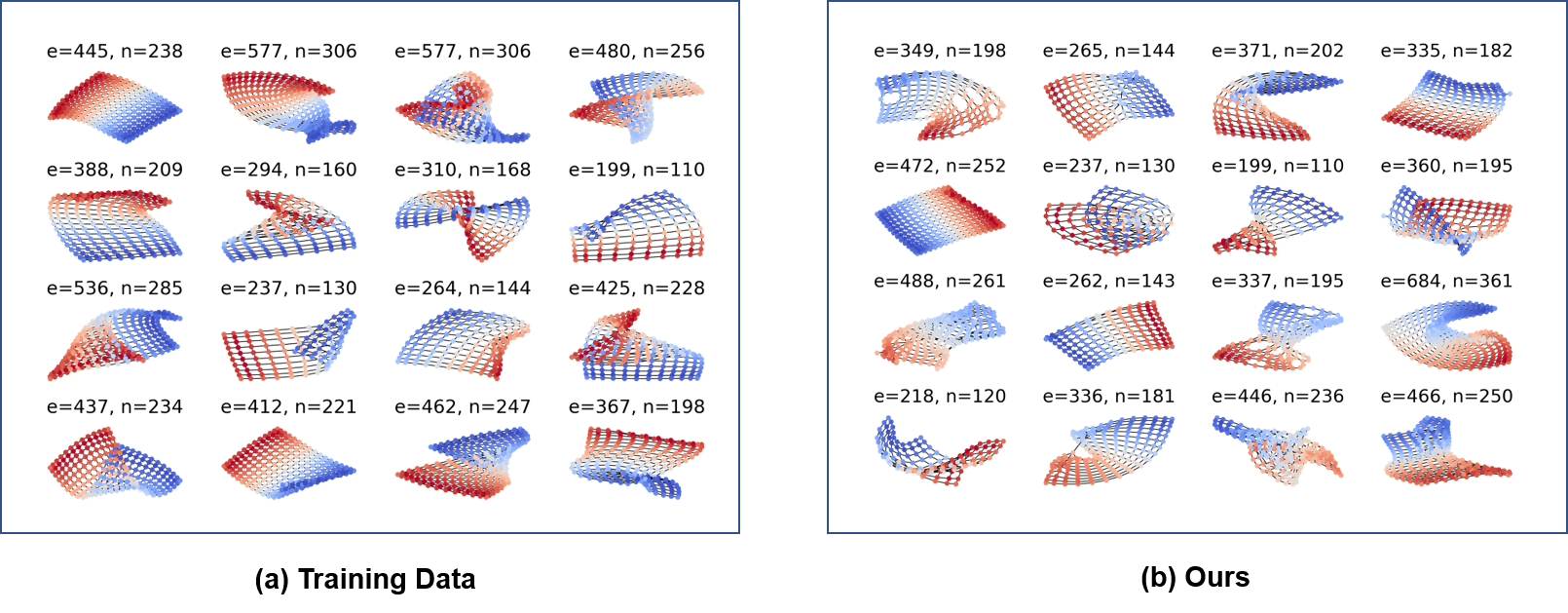}}
  \caption{Randomly selected sample Grid graphs from training set and graphs generated by SFMG.}
  \label{fig:grid-samples}
  \end{center}
  \vskip -0.2in
  \end{figure*}

\begin{figure*}[ht]
  \vskip 0.2in
  \begin{center}
  \centerline{\includegraphics[width=\textwidth]{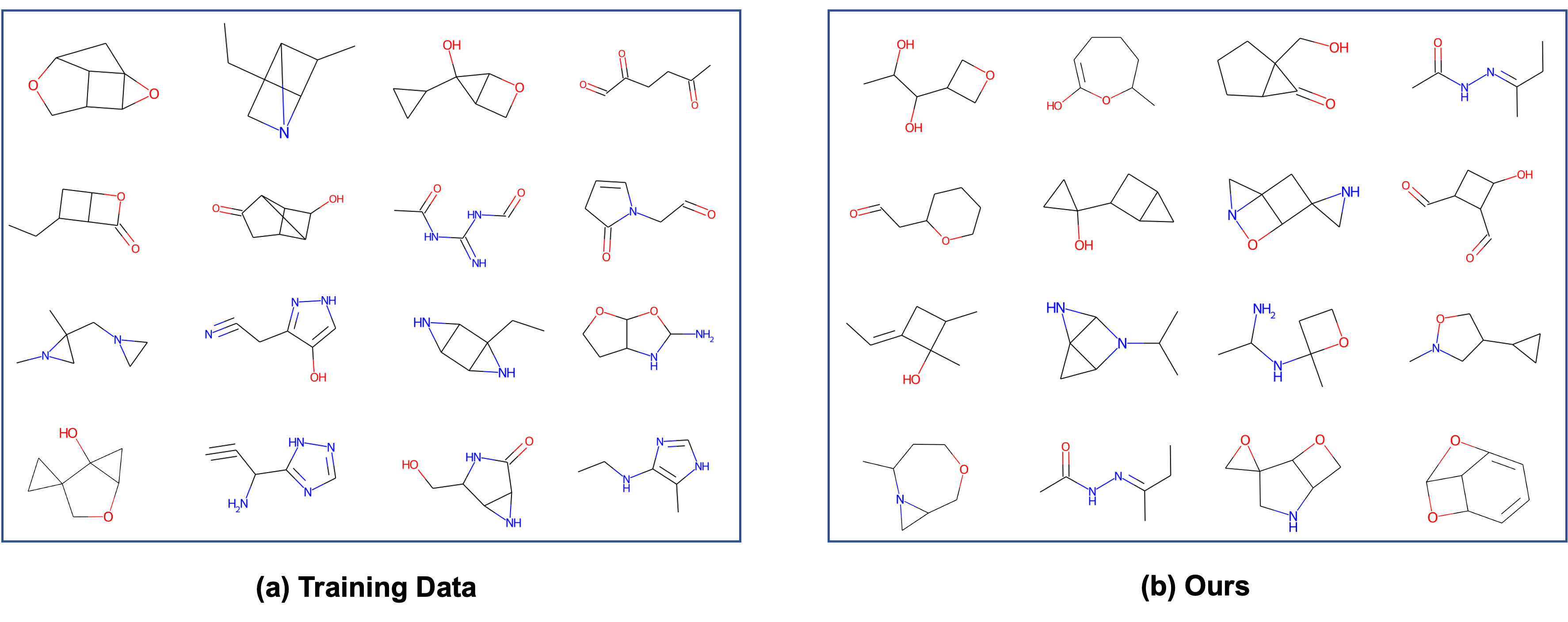}}
  \caption{Randomly selected QM9 samples from training set and graphs generated by SFMG.}
  \label{fig:qm9-samples}
  \end{center}
  \vskip -0.2in
  \end{figure*}

  In this section, we provide the visualizations of graphs from the training and generated ones by SFMG for each dataset (Figures \ref{fig:ego-samples}-\ref{fig:qm9-samples}). The visualized graphs are randomly selected from both the training and generated ones. Additionally, we provide information on the number of edges ($e$) and nodes ($n$) for general graphs.
  
  %%%%%%%%%%%%%%%%%%%%%%%%%%%%%%%%%%%%%%%%%%%%%%%%%%%%%%%%%%%%

  %%%%%%%%%%%%%%%%%%%%%%%%%%%%%%%%%%%%%%%%%%%%%%%%%%%%%%%%%%%%
  \clearpage
  \FloatBarrier

  \end{document}